\documentclass[10pt,twocolumn,letterpaper]{article}

\usepackage[pagenumbers]{cvpr} 

\definecolor{cvprblue}{rgb}{0.21,0.49,0.74}
\usepackage[pagebackref,breaklinks,colorlinks,allcolors=cvprblue]{hyperref}
\usepackage{multirow}
\usepackage{multicol}
\usepackage{booktabs} 
\usepackage[ruled]{algorithm2e}
\usepackage{wrapfig}
\usepackage{lipsum}
\usepackage{float}
\usepackage{graphicx}

\usepackage{mathtools}  

\def\paperID{10} 
\def\confName{CVPR}
\def\confYear{2026}

\title{Replay-Free Continual Low-Rank Adaptation with Dynamic Memory}

\author{Huancheng Chen\textsuperscript{1,2}, Jingtao Li\textsuperscript{1}, Weiming Zhuang\textsuperscript{1}, Chen Chen\textsuperscript{1}, Lingjuan Lyu\textsuperscript{1,*}\\
\textsuperscript{1}Sony AI, \textsuperscript{2}University of Texas at Austin \\
{\tt\small huanchengch@utexas.edu, \{jingtao.li, weiming.zhuang, ChenA.Chen,lingjuan.lv\}@sony.com}
}

\begin{document}
\maketitle
\begin{abstract}
We revisit continual learning (CL), which enables pre-trained vision transformers (ViTs) to sequentially fine-tune on new downstream tasks over time. However, as the scale of these models increases, catastrophic forgetting remains a more serious challenge. Recent studies highlight a crossover between CL techniques and parameter-efficient fine-tuning (PEFT), which focuses on fine-tuning only a small set of trainable parameters to adapt to downstream tasks, such as low-rank adaptation (LoRA). Several LoRA-based CL methods have been proposed to avoid forgetting by decoupling representations learned from previous tasks and those learned for new tasks, achieving a fixed balance between stability and plasticity, yet lacking mechanisms to dynamically adjust learned representations over time. To address this gap, we propose a novel PEFT-CL method called Dual Low-Rank Adaptation (DualLoRA), which introduces both an orthogonal LoRA adapter and a residual LoRA adapter parallel to pre-trained weights in each layer. These components are orchestrated by a dynamic memory mechanism to strike a balance between stability and plasticity. Additionally, we propose a scheme to predict task identity with confidence and calibrate the model's outputs accordingly. On ViT-based models, we demonstrate that DualLoRA offers significant advantages in accuracy, inference speed, and computation efficiency in training over existing CL methods across multiple benchmarks.
\end{abstract}

\section{Introduction}
\label{intro}
Continual learning (CL) \citep{wang2024comprehensive}, which aims to train models on a sequence of tasks, often suffers from \emph{catastrophic forgetting}—a substantial degradation in performance on previously learned tasks when adapting to new ones. This challenge persists even when continually fine-tuning vision foundation models \citep{dualprompt, l2p}, despite their strong generalization capability and robustness.\renewcommand\thefootnote{}
\footnote{*Corresponding Author: Lingjuan Lyu}
\renewcommand\thefootnote{\arabic{footnote}}Replay-based CL methods \citep{co2l} attempt to alleviate forgetting by retaining a subset of data from prior tasks as \emph{exemplars}; however, such approaches are often impractical due to storage constraints and data retention policies. Alternatively, architecture-based CL methods \citep{loo2020generalized} mitigate interference by allocating task-specific parameters and relying on task identifiers during training. Nevertheless, modifying the architecture of vision foundation models for downstream adaptation may undermine their pre-trained representations, as these weights are learned under a fixed architectural design. Moreover, many of these methods assume access to task identity at inference time—an assumption that is frequently violated in real-world scenarios.

\begin{figure}[t] 
    \centering
        \includegraphics[width=1\linewidth]
        {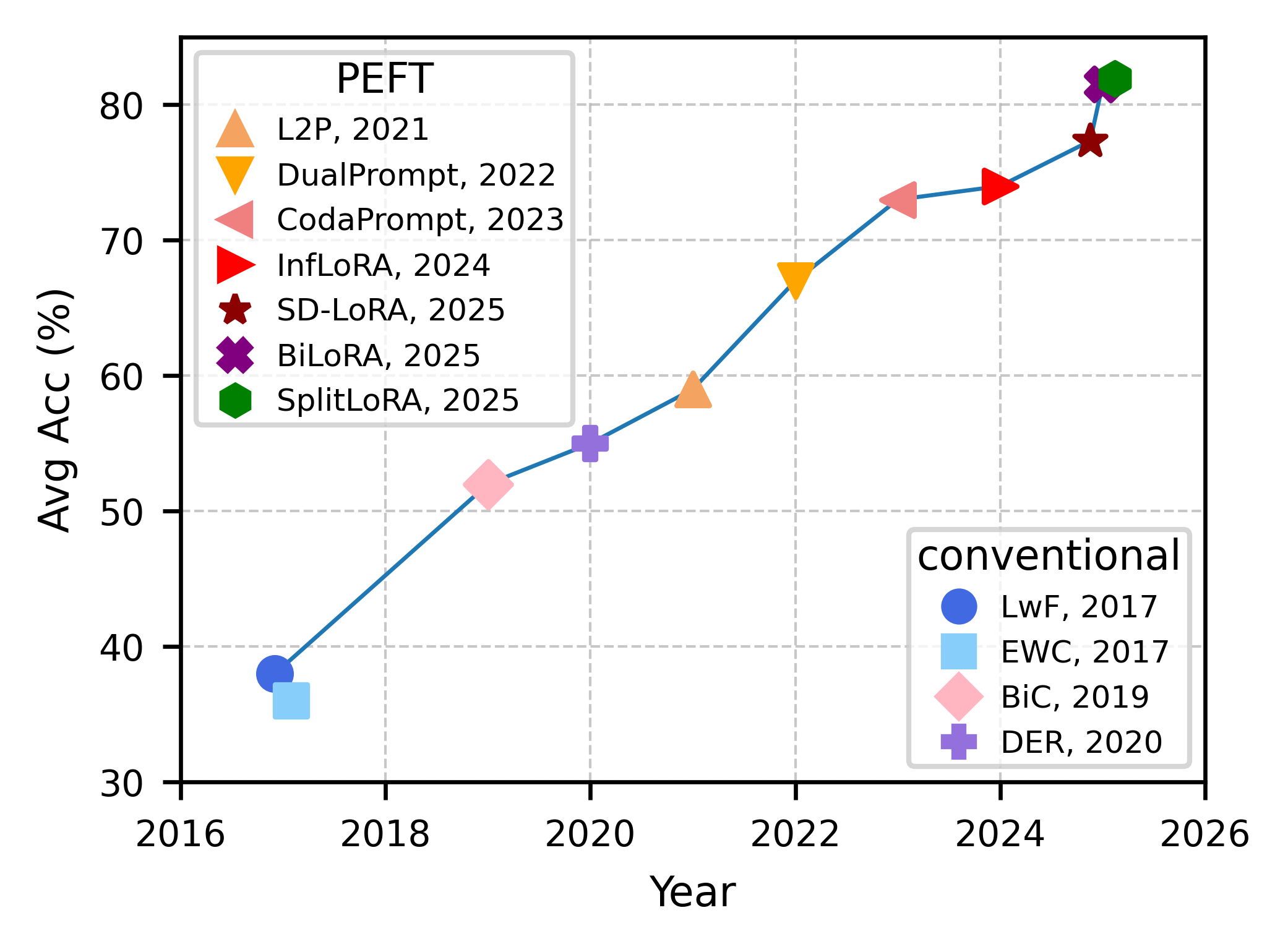}
	\caption{PEFT-based continual learning schemes dominate the
ImageNet-R dataset in recent years.}
\label{benchmark} 
\end{figure}

Recently, parameter-efficient fine-tuning (PEFT) techniques have attracted increasing attention for their ability to adapt foundation models to downstream tasks by updating or introducing only a small number of trainable parameters. Beyond computational efficiency, PEFT methods have also demonstrated notable robustness in mitigating catastrophic forgetting under sequential task adaptation. In particular, prompt-tuning \citep{dualprompt, l2p, coda} and low-rank adaptation (LoRA) \citep{inflora} have emerged as two widely adopted PEFT paradigms, achieving strong performance in continual learning and substantially outperforming conventional CL approaches on established benchmarks. Prompt-tuning methods learn a \emph{prompt pool} that matches an input image with a set of prompt vectors to align image features with patch tokens. However, existing prompt-based CL approaches rely on the original pre-trained encoder as the query function. Consequently, image tokens must be processed through the backbone network \textbf{twice}, resulting in considerable computational overhead and increased inference latency.

In contrast, LoRA-based methods typically require fewer trainable parameters to achieve comparable performance on domain-specific tasks and enable faster inference than prompt-tuning. Nevertheless, vanilla LoRA \citep{lora} struggles in continual learning settings due to severe interference across sequential tasks. Inspired by gradient projection techniques \citep{gpm}, InfLoRA \citep{inflora} takes an initial step toward addressing this issue by initializing LoRA adapters for new tasks within a subspace orthogonal to the gradient subspace of previously learned tasks. Building upon this idea, a series of follow-up studies \citep{CL-LoRA, bilora, splitlora, sdlora, keeplora} move beyond the strict orthogonality constraint imposed by InfLoRA and explore diverse forms of separated learning spaces, such as almost-orthogonal subspaces, principal–residual subspaces, and shared–specific subspaces, to better balance the stability–plasticity trade-off in continual learning. However, these LoRA-based CL methods still rely on a fixed set of adapted model weights during inference and do not incorporate dynamic adaptation of learned representations, which could further mitigate catastrophic forgetting.

To address this limitation, we propose a novel continual learning framework termed \underline{dual} \underline{lo}w-\underline{r}ank \underline{a}daptation (DualLoRA). DualLoRA integrates an orthogonal adapter and a residual adapter into each layer of pre-trained vision transformers (ViTs). Specifically, the orthogonal adapter $\mathbf{O}$ is updated exclusively along directions orthogonal to the feature subspaces extracted from previously learned tasks, while the residual adapter $\mathbf{R}$ is updated within a task-specific subspace spanned by residual bases derived from prior tasks. This design promotes \emph{stability}—by preventing interference with previously acquired knowledge through orthogonal adapters—while simultaneously enhancing \emph{plasticity} via residual adapters that facilitate efficient adaptation to new tasks. 

\begin{figure*}[t] 
\begin{center}
\includegraphics[width=1\linewidth]{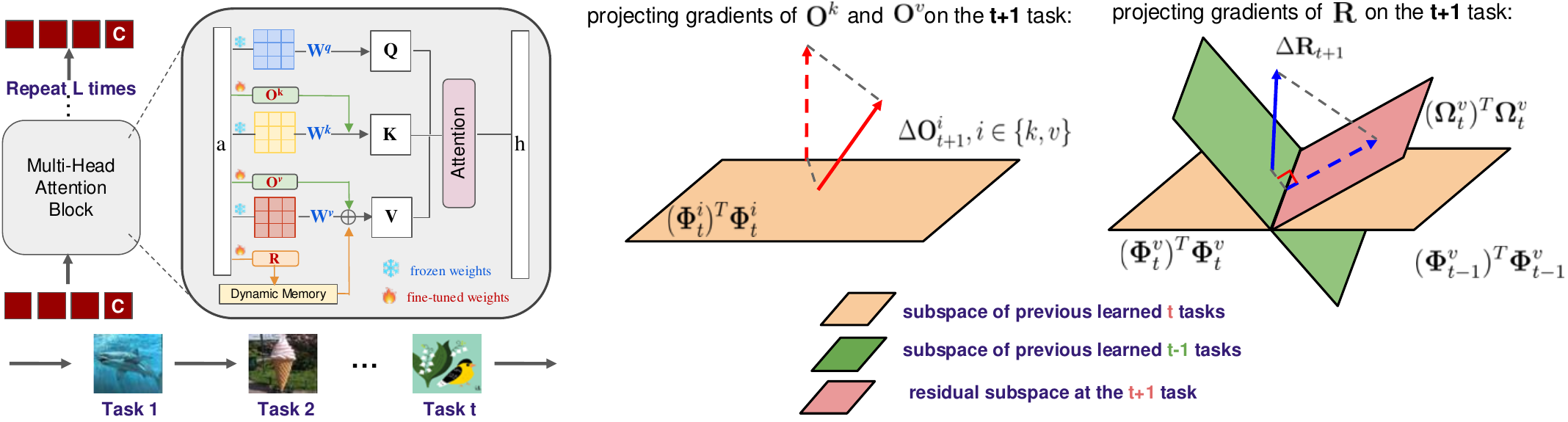}
\end{center}
\caption{Illustration of our proposed DualLoRA paradigm (left) and design insights of orthogonal adapter and residual adapter (right), where the solid arrow denotes the original update and the dashed arrow denotes the projected update. }
    \label{dual} 
\end{figure*}

DualLoRA efficiently extracts orthogonal bases from the feature subspaces of past tasks and projects the updates of $\mathbf{O}$ and $\mathbf{R}$ using matrices constructed from these bases. Moreover, these bases are further leveraged at inference time to dynamically modulate the residual adapters based on input-specific task relevance, thereby suppressing components that may degrade test performance. We refer to this inference-time mechanism as \emph{dynamic memory} (DM), as illustrated in Fig.~\ref{dual}. Extensive experimental results demonstrate that DualLoRA consistently outperforms existing PEFT-based continual learning methods across diverse benchmarks, without incurring significant additional computational or memory overhead. The main contributions of this paper are summarized as follows:
\begin{itemize}
    \item We introduce a novel low-rank adaptation paradigm for fine-tuning ViTs in continual learning settings. This paradigm efficiently extracts feature subspaces from previously learned tasks using singular value decomposition and mitigates catastrophic forgetting by reducing task interference through gradient projection.
    \item To address the challenge of limited update space due to gradient projection, we design a dual LoRA structure consisting of an orthogonal adapter and a residual adapter. This design incorporates the proposed dynamic memory mechanism, effectively balancing stability and plasticity in continual learning.
    \item To further enhance the performance of DualLoRA, we develop a simple and efficient method for inferring task identities of test samples during inference, leveraging the extracted feature subspaces. Extensive experimental results demonstrate the superior performance of DualLoRA compared to state-of-the-art baselines.
\end{itemize}

\section{Background and Related Work}
\label{related_work}
\subsection{Gradient Projection in CL}
Gradient projection is widely employed in continual learning to mitigate catastrophic forgetting by updating parameters in directions that minimize interference with previously learned tasks. OGD \citep{OGD} was the first to implement gradient descent in directions orthogonal to the stored gradient directions computed from previous tasks. The follow-up work GPM \citep{gpm} extracts orthogonal bases of task representations from randomly selected training data via singular value decomposition (SVD). A subsequent study, TRGP \citep{trgp}, introduces the concept of a trust region, allowing partial reuse of selected bases from previous tasks. FSDGPM \citep{FSGPM} further evaluates the importance of bases in GPM by assessing the sharpness of the loss landscape, assigning weights to the bases in the projection matrix according to their relative importance. Due to the computational expense of determining loss landscape sharpness, SGP \citep{sgp} offers an alternative by using accumulated singular values as an importance indicator to scale the projection matrix in GPM. Additionally, several studies \citep{NSCL,adns,dualgpm, SD, cha2025task} have explored relaxing the orthogonality constraints and optimizing the relaxation factor to enhance performance. However, these approaches are primarily developed for relatively simple models such as CNNs and face significant challenges when applied to advanced architectures like ViTs. For instance, the high dimensionality of feature embeddings in ViTs leads to substantial computational overhead when performing SVD. This motivates us to explore more efficient methods for extracting orthogonal bases from feature subspaces in ViTs.

\subsection{Parameter-Efficient Fine-Tuning in CL}
Parameter-efficient fine-tuning (PEFT) methods adapt pre-trained models to downstream tasks by fine-tuning only a small number of parameters. Among them, prompt-tuning \citep{lester2021power} has shown strong robustness in sequential learning and achieved remarkable success in continual learning benchmarks, significantly outperforming traditional continual learning schemes. The pioneering work L2P \citep{l2p} addresses continual learning by introducing a prompt pool that selects the top-$k$ relevant queries as supplementary inputs to facilitate feature alignment in the pre-trained model. DualPrompt \citep{dualprompt} further introduces task-invariant \emph{G-Prompts} and task-specific \emph{E-Prompts} to capture both shared and task-specific knowledge across tasks. S-Prompt \citep{Sprompt} learns domain-specific prompts by applying K-means clustering to training features and using a K-NN algorithm to match test samples with corresponding prompts. CodaPrompt \citep{coda} shifts from fixed instance-specific prompts to a set of prompt components, selecting a weighted combination for each sample. Moreover, several studies \citep{mcdonnell2023ranpac, gao2024beyond, zhou2024expandable} propose fine-tuning expandable prompt adapters to improve model learnability. Despite their success, these methods require forwarding training and testing samples through a query function (typically the original pre-trained model) to extract features before fine-tuning and inference, resulting in longer training and inference times.

Recently, PGP \citep{pgp} and VPT-NSP \citep{lu2024visual} adapt the idea of GPM \citep{gpm} to prompt-tuning by applying orthogonal projection to prompt gradients to mitigate forgetting, but inherit the drawbacks of prompt-tuning. Similarly, InfLoRA \citep{inflora} is the first to apply gradient projection to low-rank adaptation for ViT models by storing gradient directions from learned tasks during fine-tuning. Several follow-up studies further extend LoRA-based continual learning. CL-LoRA \citep{CL-LoRA} decomposes LoRA parameters into task-shared and task-specific components to balance stability and plasticity. SD-LoRA \citep{sdlora} introduces a scalable decoupled LoRA design to better separate task knowledge. Split-LoRA \citep{splitlora} splits the gradient space into complementary subspaces to improve stability–plasticity trade-offs. BiLoRA \citep{bilora} enforces almost-orthogonal parameter spaces across tasks to reduce interference. KeepLoRA \citep{keeplora} adapts LoRA updates via residual gradient modulation to preserve previously learned knowledge.


\section{Preliminary}
\label{preliminary}
\subsection{Continual Learning Problem Setting}
Given a pre-trained model with backbone parameters $\mathbf{W}_{0}$, we aim to fine-tune the model by adding adapters $\mathcal{A}_{l}$ in the $l$-th layer of the model and the classifier $\mathcal{F}$ to fit a sequence of domain data $\mathcal{D}_{t} = \{\mathbf{x}_{i}, \mathbf{y}_{i}\}_{i = 1}^{|\mathcal{D}_{t}|}$, where $\mathbf{x}_{i}$ denotes data samples and $\mathbf{y}_{i} \in \mathcal{Y}_{t}$ denotes corresponding labels in the $t$-th task. In the challenging class-incremental setting, there is no intersection between the label sets from two different tasks as $\mathcal{Y}_{t_{1}} \cap \mathcal{Y}_{t_{2}} = \emptyset$ for all $t_{1} \neq t_{2}$. 
When learning a new task, access to old task data might become unavailable due to storage constraints. The objective function in continual learning is to minimize the empirical risk of the unified adapters $\mathcal{A}_{1:L}$ integrated in $L$ layers of the pre-trained model that performs inference on the sequence of data from $T$ various tasks as follows:
\begin{equation}
    \min_{\{\mathcal{A}_{l}\}_{l=1}^{L}, \mathcal{F}} \frac{1}{T}\sum_{t=1}^{T} \mathcal{L}_{\text{task}}(\mathbf{W}_{0}, \mathcal{A}_{1:L}, \mathcal{F}, \mathcal{D}_{t}),
\label{eq:cl_objective}
\end{equation}
where $\mathbf{W}_{0}$ is the frozen backbone parameters of the pre-trained model; $\mathcal{L}_{\text{task}}(\cdot)$ is the loss function depending on the specific task. The classifier $\mathcal{F}$ consists of an expanding set of fully-connected layers $f_{t}: \mathbb{R}^{d} \xrightarrow{} \mathbb{R}^{C_{t}}$, where $d$ is the dimension of embedding and $C_{t}$ is the number of classes in the $t$-th task. Given \textbf{no task identities} during inference, all the learned $f_{t}(\cdot)$ are used to predict categories of input data.

\subsection{Multi-Head Attention Block}
Vision transformer (ViT) models break down images into $n$ patches and flatten these patches into patch embeddings of dimension $d$. The encoder of a ViT consists of a sequence of multi-head attention (MHA) blocks containing \emph{key}, \emph{query} and \emph{value} weights $\mathbf{W}^{q}, \mathbf{W}^{k}$ and $\mathbf{W}^{v}$ for mapping input activation signals $\mathbf{a}^{(l)}$ into $\mathbf{Q}^{(l)}$, $\mathbf{K}^{(l)}$ and $\mathbf{V}^{(l)}$ and obtaining the output signals $\mathbf{h}^{(l)}$ by computing
\begin{equation}
\label{qkv}
    \mathbf{h}^{(l)} = \text{softmax} \left( \frac{\mathbf{Q}^{(l)}\left(\mathbf{K}^{(l)}\right)^{\top}}{\sqrt{d}} \right) \cdot \mathbf{V}^{(l)},
\end{equation}
where $\mathbf{Q}^{(l)} := \mathbf{a}^{(l)}\mathbf{W}^{q}$, $\mathbf{K}^{(l)} :=  \mathbf{a}^{(l)}\mathbf{W}^{k}$ and $\mathbf{V}^{(l)} := \mathbf{a}^{(l)}\mathbf{W}^{v}$. The output signals $\mathbf{h}^{(l)}$ are input to the feed-forward network and passed through normalization before being forwarded to the next MHA block until $\mathbf{h}^{(L)}$ is directed to the classifier.

\subsection{Low-Rank Adaptation}
\label{pre_lora}
Low-rank adaptation (LoRA) is a parameter-efficient fine-tuning method that enables reducing memory consumption by assigning learnable low-rank matrices $\mathbf{A}\in \mathbb{R}^{r \times d}$ and $\mathbf{B}\in \mathbb{R}^{d \times r}$ parallel to the frozen pre-trained weights $ \mathbf{W}_{0} \in \mathbb{R}^{d \times d}$ into each layer of the model as follows,
\begin{equation}
    \mathbf{W} := \mathbf{W}_{0} + \mathbf{B} \mathbf{A},
\end{equation}
where $ \mathbf{W}$ denotes model weights after fine-tuning, and $r \ll d$. In this paper, we follow the strategy in \citep{gao2023unified}, and only implement LoRA fine-tuning on $\mathbf{W}_{0}^{k}$ and $\mathbf{W}_{0}^{v}$ while keeping $\mathbf{W}_{0}^{q}$ frozen during the whole procedure.

\section{Methodology: Dual Low-Rank Adaptation}
\label{method}
The subspace for updating the model becomes more constrained as the gradients are projected into a subspace orthogonal to all feature subspaces from the previous tasks. 
Subsequent studies \citep{FSGPM, NSCL, trgp, SD} ease the stringent constraints of orthogonality to expand the optimization subspace in a new task, considering the stability-plasticity trade-off. 
Inspired by the prior studies, we propose a novel low-rank adaptation structure, DualLoRA, consisting of an orthogonal adapter $\mathbf{O} := \mathbf{A}_{o}\mathbf{B}_{o} \in \mathbb{R}^{d \times d}$ and a residual adapter $\mathbf{R} :=
\mathbf{A}_{r}\mathbf{B}_{r} \in \mathbb{R}^{d \times d}$ that are updated in the orthogonal direction and residual direction, as shown in Fig~\ref{dual}. 
We follow the strategy in the existing PEFT continual learning schemes \citep{dualprompt, l2p, pgp, coda, inflora}, using pre-trained vision transformers (ViTs) as backbone models throughout this paper. In the forthcoming sections, we will illustrate the process of updating both adapters and integrating dynamic memory during model inference.

\subsection{Orthogonal Adapter}
The milestone work GPM \citep{gpm}, which uses orthogonal gradient projection to mitigate forgetting, involves flattening feature maps extracted by convolutional kernels into vectors and performing singular value decomposition (SVD) on these vectors to obtain orthogonal feature bases. However, vectorizing the patch embeddings of a ViT with dimensions $(n,d)$ necessitates extensive computation in SVD, particularly when dealing with high-resolution inputs. In vision transformers, feature embeddings are often redundant for classification tasks, as only the first embedding (commonly referred to as the \emph{class token}) is passed to the classifier for prediction.

To this end, we propose an efficient method for extracting the orthogonal bases of the class-token subspace without performing SVD on the entire high-dimensional embedding space. Specifically, given the pre-trained weights $\mathbf{W}_{0}^{q}$, $\mathbf{W}_{0}^{k}$, and $\mathbf{W}_{0}^{v}$, the fine-tuned \emph{key} and \emph{value} weights, i.e., $\mathbf{W}_{t+1}^{k}$ and $\mathbf{W}_{t+1}^{v}$, for the $(t+1)$-th task can be derived as follows:
\begin{equation}
    \label{eq:wk_orth_update}
    \mathbf{W}_{t+1}^{k} = \mathbf{W}_{0}^{k} + \sum_{\tau = 1}^{t+1} \Delta \mathbf{O}_{\tau}^{k} = \mathbf{W}_{t}^{k} + \Delta \mathbf{O}_{t+1}^{k},
\end{equation}
\begin{equation}
    \label{eq:wv_orth_update}
    \mathbf{W}_{t+1}^{v} = \mathbf{W}_{0}^{v} + \sum_{\tau = 1}^{t+1} \Delta \mathbf{O}_{\tau}^{v} = \mathbf{W}_{t}^{v} + \Delta \mathbf{O}_{t+1}^{v},
\end{equation}
where $\Delta \mathbf{O}_{\tau}^{k}, \Delta \mathbf{O}_{\tau}^{v}$ are the updates of the orthogonal adapter computed from the $\tau$-th task.
According to \eqref{qkv}, when we fine-tune the parameters on the $(t+1)$-th task, the change of output signal $\mathbf{h}^{(l)}$ given the same data can be formulated as
\begin{equation}
\begin{aligned}
    \label{update_diff}
     \Delta \mathbf{h}^{(l)} \approx & \quad  \Gamma \cdot \frac{\mathbf{Q}^{(l)} \left(\mathbf{a}^{(l)} \Delta \mathbf{O}_{t+1}^{k} \right)^{\top}}{\sqrt{d}}  \cdot \mathbf{V}^{(l)} \\
     & + \underbrace{\text{softmax} \left( \frac{\mathbf{Q}^{(l)} \left(\mathbf{K}^{(l)}\right)^{\top}}{\sqrt{d}} \right) \cdot \mathbf{a}^{(l)}}_{\mathbf{S}^{(l)}} \Delta \mathbf{O}_{t+1}^{v},
\end{aligned}
\end{equation}
where $\Gamma$ is a diagonal matrix (the derivation is deferred to Appendix A.1).
To preserve the value of the class token in $\mathbf{h}^{(L)}$ output by the last layer, we must restrict the value of $\Delta \mathbf{h}^{(l)}_{1} \approx \mathbf{0}$ (the change of the first row in $\mathbf{h}^{(l)}$) for each layer so that the class token of the same test sample from old tasks can be preserved after fine-tuning on the new task. Since $\mathbf{Q}^{(l)}$ is unchanged with the frozen weight $\mathbf{W}_{0}^{q}$, we need to project $\Delta \mathbf{O}_{t+1}^{k}$ into the subspace orthogonal to the subspace spanned by $\mathbf{k}^{(l)} := \mathbf{Q}_{1}^{(l)}$, denoting the first row of $\mathbf{Q}^{(l)}$. Meanwhile, $\Delta \mathbf{O}_{t+1}^{v}$ must be orthogonal to the subspace of $\mathbf{v}^{(l)} := \mathbf{S}_{1}^{(l)}$, the first row of $\mathbf{S}^{(l)}$. Following the strategy in GPM \citep{gpm}, we randomly sample $m$ data points from the current task after fine-tuning on the $t$-th task and input these $m$ samples to the model for obtaining embedding matrices $\tilde{\mathbf{K}}^{(l)} \in \mathbb{R}^{m \times d}$ consisting of $\{\mathbf{k}_{i}^{(l)}\}_{i=1}^{m}$ and $\tilde{\mathbf{V}}^{(l)} \in \mathbb{R}^{m \times d}$ consisting of $\{\mathbf{v}_{i}^{(l)}\}_{i=1}^{m}$. We update the new feature matrices $\boldsymbol{\Phi}_{t}^{k}$ and $\boldsymbol{\Phi}_{t}^{v}$ by extracting the orthogonal bases of $\tilde{\mathbf{K}}^{(l)}$ and $\tilde{\mathbf{V}}^{(l)}$ using SVD and concatenating them into the previous feature matrices $\boldsymbol{\Phi}_{t-1}^{k}$ and $\boldsymbol{\Phi}_{t-1}^{v}$. With the feature matrices obtained on the $t$-th task, we can project the updates $\Delta \mathbf{O}_{t+1}^{k}$ and $\Delta \mathbf{O}_{t+1}^{v}$ by
\begin{equation}
\label{projection}
    \Delta \mathbf{O}_{t+1}^{i} \xleftarrow{} \Delta \mathbf{O}_{t+1}^{i} - (\boldsymbol{\Phi}_{t}^{i})^{\top}\boldsymbol{\Phi}_{t}^{i} \Delta\mathbf{O}_{t+1}^{i},\quad \forall i \in \{k, v\}.
\end{equation}
When we select $m$ data points, with $m \ll d$, for extracting orthogonal bases, the complexity of SVD is $\mathcal{O}(m^{2}d)$, which is much more efficient than SVD with $\mathcal{O}(d^{3})$ implemented in the previous work InfLoRA \citep{inflora}. We emphasize that our orthogonal adapters undergo a different update process than InfLoRA, where orthogonal bases are extracted from the gradient subspace. Instead, we develop an alternative feature set, $\mathbf{S}^{(l)}$, specifically to preserve the class token, as outlined in \eqref{update_diff}. Since we only extract the orthogonal subspace from $\Delta \mathbf{h}_{1}^{(l)}$, we cannot guarantee that $\Delta \mathbf{h}_{1}^{(l)} = \mathbf{0}$ in every layer. However, reducing the magnitude of $\Delta \mathbf{h}_{1}^{(l)}$ can mitigate catastrophic forgetting, even if the value is not zero (details are provided in Appendix A.2).

\subsection{Residual Adapter}
\label{residual}
As the feature subspaces represented by $\boldsymbol{\Phi}_{t}^{k}$ and $\boldsymbol{\Phi}_{t}^{v}$ expand with the accumulation of learned tasks, the majority of the components in the updates, $\Delta \mathbf{O}_{t+1}^{k}$ and $\Delta \mathbf{O}_{t+1}^{v}$, are progressively subtracted, as detailed in \eqref{projection}. Consequently, the update magnitudes approach zero, resulting in diminished performance during fine-tuning on new tasks. To address this issue, we introduce a residual adapter $\mathbf{R}_{t+1}$ (initialized as $\mathbf{0}$) in parallel with $\mathbf{O}_{t+1}^{v}$, providing extra capacity for new tasks and maintaining a balance between stability and plasticity.

When the model is fine-tuned on the $(t+1)$-th task, the residual adapter $\mathbf{R}_{t+1}$ is updated within the subspace $\mathcal{R}_{t}$ spanned by $\boldsymbol{\Psi}_{t}$ (we set $\boldsymbol{\Psi}_{1} = \emptyset$) defined as
\begin{equation}
    \label{knowledge}
    \boldsymbol{\Psi}_{t} := \boldsymbol{\Phi}_{t}^{v} \setminus \boldsymbol{\Phi}_{t-1}^{v} \subseteq \mathbb{R}^{d},
\end{equation}
where $\boldsymbol{\Phi}_{t}^{v}$ and $\boldsymbol{\Phi}_{t-1}^{v}$ are obtained after fine-tuning on the $t$ and $t-1$ tasks. It is worth noting that the subspaces $\mathcal{R}_{t}$ and $\mathcal{R}_{t+1}$ are specific to their corresponding tasks $t$ and $t+1$, respectively, as $\boldsymbol{\Psi}_{t+1} \cap \boldsymbol{\Psi}_{t} = \emptyset$. Specifically, the subspace $\mathcal{R}_{t}$ indicates the residual knowledge extracted from the most recent task, providing supplementary bases to enlarge the optimization subspace. With these extracted bases, we are able to project the updates $\Delta\mathbf{R}_{t+1}$ into the subspace $\mathcal{R}_{t}$ by 
\begin{equation}
  \Delta \mathbf{R}_{t+1} \xleftarrow{} \boldsymbol{\Psi}_{t}^{\top} \boldsymbol{\Psi}_{t} \Delta\mathbf{R}_{t+1}.
\end{equation}

When we conduct \textbf{fine-tuning} on the $(t+1)$-th task, the value matrix $\mathbf{V}^{(l)}$ in the $l$-th layer given the input activations $\mathbf{a}^{(l)}$ can be found as
\begin{equation}
     \mathbf{V}^{(l)} = \mathbf{a}^{(l)} \left(\mathbf{W}_{0}^{v} +  \mathbf{O}_{t+1}^{v}\right) + \mathbf{a}^{(l)}\mathbf{R}_{t+1} =  \mathbf{V}_{o}^{(l)} + \mathbf{V}_{r}^{(l)},
\end{equation}
where $\mathbf{W}_{0}^{v}$ denotes the pre-trained weights, and $\mathbf{V}_{o}^{(l)} := \mathbf{a}^{(l)} \left(\mathbf{W}_{0}^{v} +  \mathbf{O}_{t+1}^{v}\right)$, $\mathbf{V}_{r}^{(l)} := \mathbf{a}^{(l)} \mathbf{R}_{t+1}$.

\begin{table*}[t]
\caption{Metrics (\%) computed from experiments on ImageNet-R. We report the average accuracy over 3 trials, each with different random seeds. The numeric after "$\pm$" denotes standard deviation.} 
\centering  
\begin{tabular}{lcccccc}
\bottomrule[1pt]
\label{table1}
\multirow{2}{*}{\textbf{Method} }    & 
 \multicolumn{2}{c}{\textbf{5-Split ImageNet-R}} & \multicolumn{2}{c}{\textbf{10-Split ImageNet-R}} & \multicolumn{2}{c}{\textbf{20-Split ImageNet-R}}  \\
   &    ACC($\uparrow$) & FT($\downarrow$) & ACC($\uparrow$) & FT($\downarrow$)  & ACC($\uparrow$) & FT($\downarrow$)  \\

\hline  
LoRA   &72.33 $\pm 0.94$&12.1 $\pm 1.19$&61.85 $\pm 0.52$& 26.0 $\pm 1.35$ &48.59 $\pm 0.39$ &34.4 $\pm 0.57$ \\
L2P   &61.60 $\pm 0.43$ &5.36 $\pm 0.27$ &59.21 $\pm 0.68$ &7.59 $\pm 0.78$ &56.36 $\pm 0.83$ &10.3 $\pm 0.72$\\
DualPrompt   &68.47 $\pm 0.23$ & 3.18 $\pm 0.24$ &66.72  $\pm 0.30$&4.15 $\pm 0.11$ &64.40 $\pm 0.18$ & 5.82 $\pm 0.51$ \\
PGP   &69.07 $\pm 0.28$ &3.41 $\pm 0.18$ &64.22 $\pm 4.53$  &4.23 $\pm 0.22$  & 64.19 $\pm 0.38$ & 6.50 $\pm 0.31$  \\
S-Prompt  &51.33 $\pm 0.22$ &27.6 $\pm 1.18$ & 49.80 $\pm 0.16$ &29.2 $\pm 0.93$ &55.64 $\pm 0.53$  &22.3 $\pm 1.85$\\
CodaPrompt  &74.91 $\pm 0.30$ &1.85 $\pm 0.07$ & 73.83 $\pm 0.29$&2.56 $\pm 0.31$&68.96 $\pm 0.46$ & \textbf{3.25 }$\pm 0.40$\\
InfLoRA  &77.30 $\pm 0.49$ &3.05 $\pm 0.44$ & 74.03 $\pm 0.30$  &6.18 $\pm 0.25$ &69.77 $\pm 0.31$  &7.98 $\pm 0.40$ \\
\hline 
DualLoRA   &78.55 $ \pm 0.12$ &2.61 $\pm 0.25$  &76.23 $ \pm 0.33$   & 3.67 $\pm 0.66$ &71.25 $\pm 0.31$&5.45 $\pm 0.27$  \\
DualLoRA+  &\textbf{79.88} $\pm 0.50$& \textbf{1.10} $\pm 0.16$   &\textbf{81.17} $\pm 0.23$ &\textbf{2.04} $\pm 0.05$& \textbf{74.73} $\pm 0.40$ &3.75 $\pm 0.10$\\
\toprule[1pt]
\end{tabular}
\end{table*}
\begin{table*}[t]
\caption{Metrics (\%) computed from experiments on CIFAR100 and Tiny-ImageNet. We report the average accuracy over 3 trials, each with different random seeds.  The numeric after "$\pm$" denotes standard deviation.} 
\centering
\begin{tabular}{lcccccc}
\bottomrule[1pt]
\label{table2}
\multirow{2}{*}{\textbf{Method} }   & 
 \multicolumn{2}{c}{\textbf{10-Split CIFAR100}} & \multicolumn{2}{c}{\textbf{10-Split TinyImageNet}} & \multicolumn{2}{c}{\textbf{20-Split TinyImageNet}}  \\
   &     ACC($\uparrow$) & FT($\downarrow$) & ACC($\uparrow$) & FT($\downarrow$)  & ACC($\uparrow$) & FT($\downarrow$)\\
\hline  
LoRA  &  73.32 $\pm 0.38$ & 20.4 $\pm 0.53$ &67.69 $\pm 0.49$ & 23.7 $\pm 0.65$ &48.48 $\pm 2.36$ & 44.4 $\pm 2.73$ \\
L2P &83.97 $\pm 0.18$ &6.41 $\pm 0.09$  &81.90 $\pm 0.42$ &5.39 $\pm 0.33$  & 81.24 $\pm 0.21$ &5.86 $\pm 0.22$  \\
DualPrompt  &85.85 $\pm 0.22$& 5.41 $\pm 0.12$  &85.10 $\pm 0.10$ &3.95 $\pm 0.22$  &82.77  $\pm 0.12$&5.31 $\pm 0.10$   \\
PGP   &85.28 $\pm 0.01$&5.60 $\pm 0.34$  &84.83 $\pm 0.21$  &4.32 $\pm 0.16$  &83.49 $\pm 0.35$ & 5.24 $\pm 0.31$  \\
S-Prompt   &67.03 $\pm 0.66$&24.8 $\pm 0.62$ &68.41 $\pm 0.26$ &10.41$\pm 0.68$ &74.69 $\pm 0.30$ &7.70 $\pm 0.28$\\
CodaPrompt  & 85.77 $\pm 0.69$ &4.07 $\pm 0.22$ & 85.67 $\pm 0.25$ &3.16 $\pm 0.17$ &83.61 $\pm 0.47$ &\textbf{3.34}$\pm 0.35$\\
InfLoRA   &85.62 $\pm 0.74$ &4.34 $\pm 0.06$ & 81.28 $\pm 0.40$ & 8.62 $\pm 0.36$ &75.89 $\pm 0.38$ & 13.8 $\pm 0.11$\\
\hline 
DualLoRA  &89.13  $\pm 0.17$ &4.08 $\pm 0.16$&  86.42 $\pm 0.07$   &3.87 $\pm 0.18$  &83.75 $\pm 0.25$& 5.24 $\pm 0.15$ \\
DualLoRA+  &\textbf{90.94} $\pm 0.15$ &\textbf{3.20} $\pm 0.18$ & \textbf{87.74} $\pm 0.21$  & \textbf{2.45} $\pm 0.25$ & \textbf{84.65} $\pm 0.07$ & 3.61 $\pm 0.13$   \\
\toprule[1pt]
\end{tabular}
\end{table*}

\subsection{Dynamic Memory}
As previously mentioned, the residual adapter $\mathbf{R}_{t+1}$ is updated within the subspace $\mathcal{R}_{t}$, a subset of the feature subspace spanned by $\boldsymbol{\Phi}_{t}$ extracted from the $t$-th task. Consequently, the fine-tuning process may deteriorate the performance on prior tasks. To mitigate this issue, we introduce a \emph{dynamic memory} mechanism that adjusts the value of $\mathbf{V}_{r}^{(l)}$, which is the output of $\mathbf{R}_{t+1}$ during \textbf{inference} on test data according to
\begin{equation}
     \hat{\mathbf{V}}^{(l)} = \mathbf{V}_{o}^{(l)} + \mathbf{a}^{(l)} \boldsymbol{\Omega}_{t+1}^{\top}\boldsymbol{\Omega}_{t+1}\mathbf{R}_{t+1} = \mathbf{V}_{o}^{(l)} + \hat{\mathbf{V}}_{r}^{(l)},
\end{equation}
where $\boldsymbol{\Omega}_{t+1}^{\top}\boldsymbol{\Omega}_{t+1}$ is computed according to the input activation signal $\mathbf{a}^{(l)}$. Specifically, the attention score $\mathbf{S}^{(l)} \propto \text{softmax}\!\left(\frac{\mathbf{Q}^{(l)}(\mathbf{K}^{(l)})^{\top}}{\sqrt{d}}\right)$, computed by applying $\mathbf{a}^{(l)}$ to the query and key weight matrices, reflects the relevance between the input test sample and the task associated with the extracted bases used to update the residual adapter $\mathbf{R}_{t+1}$. We utilize the matrices $\boldsymbol{\Psi}_{\tau} \in \mathbb{R}^{r_{\tau} \times d}$ ($\tau \leq t+1$), stored in memory, to multiply the first row of the attention score, $\mathbf{v}^{(l)} := \mathbf{S}_{1}^{(l)}$, as follows:
\begin{equation}
\label{relevant}
    \text{if } r_{\tau} \neq 0,\;
    \omega_{\tau} =
    \frac{\left\|\boldsymbol{\Psi}_{\tau}\mathbf{v}^{(l)}\right\|}
    {r_{\tau}\left\|\mathbf{v}^{(l)}\right\|},
    \quad
    \text{otherwise, } \omega_{\tau}=0,
\end{equation}
where $r_{\tau}$ indicates the rank of $\boldsymbol{\Psi}_{\tau}$. Since the task-specific residual bases remain independent across different tasks, the cosine similarity between feature vectors $\mathbf{v}^{(l)}$ extracted from input test samples and the stored residual bases $\boldsymbol{\Psi}_{\tau}$ for each task can be used as a scaling factor for the corresponding components in the outputs of the residual adapter. This method assigns lower weights to components irrelevant to the current test samples, while components with high relevance to the samples are given proportionally higher weights. The resulting matrix $\boldsymbol{\Omega}_{t+1}$ is obtained by
\begin{equation}
    \boldsymbol{\Omega}_{t+1} =
    \left[
	\begin{array}{cccc}
	\boldsymbol{\Sigma}_{1} & \mathbf{0} & \cdots & \mathbf{0}\\
	\mathbf{0} & \boldsymbol{\Sigma}_{2} & \cdots & \mathbf{0}\\
	\vdots & & \ddots & \vdots\\
	\mathbf{0} & \mathbf{0} & \cdots & \boldsymbol{\Sigma}_{t+1}
	\end{array}
	\right]
    \left[
    \begin{array}{c}
	\boldsymbol{\Psi}_{1}\\
	\boldsymbol{\Psi}_{2}\\
	\vdots\\
	\boldsymbol{\Psi}_{t+1}
	\end{array}
	\right]
   \in \mathbb{R}^{r' \times d},
\end{equation}
where $r' = \sum_{\tau=1}^{t+1} r_{\tau}$ and $\boldsymbol{\Sigma}_{\tau} \in \mathbb{R}^{r_{\tau} \times r_{\tau}}$ is a diagonal matrix with identical entries $\omega_{\tau}^{1/2}$.

\subsection{Task Identification with Confidence}
\label{alignment}

As the number of fully connected layers $f_{t}(\cdot)$ increases with the addition of tasks during continual learning, there is a risk that an irrelevant fully connected layer may generate the maximal logit, resulting in incorrect predictions for input test samples. This motivates us to propose a task identity prediction scheme based on task relevance, computed using $\boldsymbol{\Psi}_{\tau}$ as described earlier in \eqref{relevant}. 

As mentioned earlier, we sample $m$ training data points to extract orthogonal bases after completing the $t$-th task. During this process, we obtain the average feature vector $\bar{\mathbf{v}}^{(L)}$ forwarded to the \textbf{final} attention block from each task and compute the similarity vector $\boldsymbol{\pi}_{t} = \{\omega_{\tau}\}_{\tau=1}^{t}$ based on \eqref{relevant}. Therefore, we obtain a set $\boldsymbol{\Pi}_{t} = \{\boldsymbol{\pi}_{1}, \dots, \boldsymbol{\pi}_{t}\}$ that can be used to distinguish the task identity during inference. Specifically, let $\boldsymbol{\pi}^{\star}$ denote the similarity vector computed from the input test sample; then we can predict the \textbf{task identity} by
\begin{equation}
    \hat{k} = \arg\max_{\tau} g(\boldsymbol{\pi}_{\tau},\boldsymbol{\pi}^{\star}),
\end{equation}
\begin{equation}
g(\boldsymbol{\pi}_{\tau},\boldsymbol{\pi}^{\star}) :=
    \frac{\left\|\boldsymbol{\pi}_{\tau} \cdot \boldsymbol{\pi}^{\star}\right\|}
    {\left\|\boldsymbol{\pi}_{\tau}\right\| \cdot \left\|\boldsymbol{\pi}^{\star}\right\|},
\end{equation}
\begin{equation}
    \hat{\delta} = \lambda \left(
    g(\boldsymbol{\pi}_{\hat{k}},\boldsymbol{\pi}^{\star})
    -
    \max_{\tau \neq \hat{k}} g(\boldsymbol{\pi}_{\tau},\boldsymbol{\pi}^{\star})
    \right),
\end{equation}
where $\lambda$ is a scaling factor, $\hat{k}$ denotes the predicted task identity, and $\hat{\delta}$ indicates the confidence of this prediction. Moreover, we scale the output logits as 
\begin{equation}
    f_{\hat{k}}(\mathbf{h}^{(L)}) \xleftarrow{} (1 + \hat{\delta}) \cdot f_{\hat{k}}(\mathbf{h}^{(L)}).
\end{equation}

\section{Experiments}
\label{exp}
\subsection{Experimental Settings}
\textbf{Datasets and Metrics.} We evaluate the proposed method DualLoRA on three continual learning benchmark datasets: CIFAR100, Tiny-ImageNet and ImageNet-R\citep{imagenetr}. To generate a sequence of tasks in a class-incremental setting as illustrated in \eqref{eq:cl_objective}, we randomly split the original dataset by class ID. This process creates multiple partitions, each containing an equal number of classes. Each partition corresponds to a distinct task.
Following the strategy in existing studies in continual learning \citep{pgp,coda, dualprompt}, we compute the final average accuracy (denoted by ACC) and degree of forgetting (denoted by FT) for evaluating the performance of our method, and these two metrics can be found as
\begin{equation}
    \text{ACC} = \frac{1}{T}\sum_{\tau = 1}^{T} \text{acc}_{\tau, T},
\end{equation}
\begin{equation}
    \text{FT} = \frac{1}{T-1}\sum_{\tau = 1}^{T-1} \text{acc}_{\tau, \text{best}} - \text{acc}_{\tau, T},
\end{equation}
where $\text{acc}_{\tau, T}$ denotes the accuracy of the $\tau$-th task after the model learns the $T$-th task while $\text{acc}_{\tau, \text{best}}$ denotes the  highest accuracy on the $\tau$-th task during the whole fine-tuning process. Throughout the evaluation, we assume that the task identities of the testing data are \textbf{unknown}.
\\

\noindent \textbf{Baselines.} Our baselines include vanilla LoRA, L2P \citep{l2p}, DualPrompt \citep{dualprompt}, PGP \citep{pgp}, S-Prompt \citep{Sprompt}, CodaPrompt \citep{coda}, and InfLoRA \citep{inflora}. 
We focus on comparing the proposed DualLoRA with the state-of-the-art PEFT-based CL schemes since they are superior to traditional CL schemes. To demonstrate the  \textbf{upper-bound} performance of DualLoRA, we implemented it under the setting where all samples in a batch share the same task identity, referred to as \textbf{DualLoRA+}. In this scenario, we use average feature to compute similarity in the task identity prediction as described in Section \ref{alignment} and facilitate more accurate task prediction, as the average feature exhibits less variance and is closer to the true mean.
\\

\noindent \textbf{Model Architecture and Hyperparameters.} We use ViT-B/16 backbone pretrained on ImageNet-21K as the foundation model throughout  all experiments. We use the Adam optimizer with parameters $\beta_1 = 0.9$ and $\beta_2 = 0.999$ for model fine-tuning in $5$ epochs and the batchsize is set to $16$ in all experiments. More details of hyperparameters are provided in Appendix B.1.

\begin{figure}[t] 
    \centering
	  \subfloat[ImageNet-R]{
       \includegraphics[width=0.48\linewidth]
       {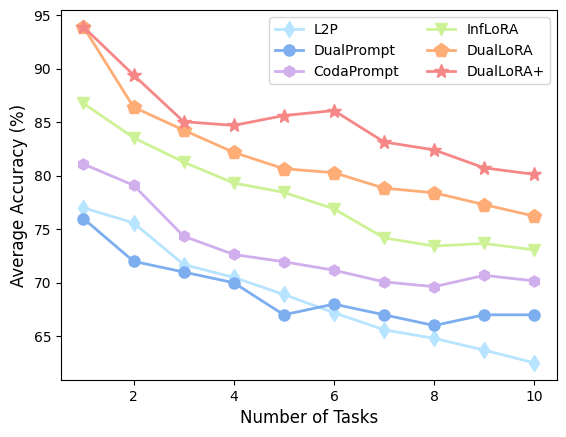}}
	  \subfloat[CIFAR100]{
        \includegraphics[width=0.5\linewidth]{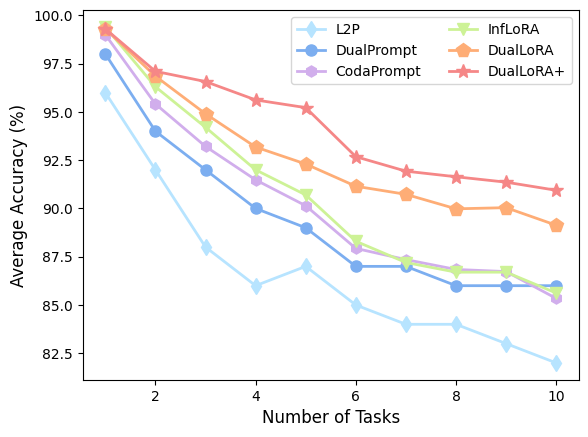}}
	\caption{Figures (a) and (b) demonstrate the average accuracy of different methods during training.}
\label{acc_vs_time} 
\end{figure}

\subsection{Experimental Results}
\textbf{ImageNet-R.} As shown in Table~\ref{table1}, DualLoRA demonstrates comparable performance with InfLoRA, which outperforms other Prompt-based CL schemes in final average accuracy ACC. DualLoRA exhibits a slight performance decrease compared to InfLoRA by $1.25\%$ on the 5-split benchmark, yet it outperforms InfLoRA by $2.2\%$ and $1.48\%$ on the 10-split and 20-split benchmarks, respectively. Forgetting serves as another metric to quantify the performance degradation on previous tasks, which may not consistently align with the average accuracy. CodaPrompt shows superior performance in mitigating forgetting, even though it does not achieve the same level as InfLoRA and DualLoRA in the average accuracy metric. DualLoRA+ significantly enhances both average accuracy and forgetting metrics, surpassing all other baselines except for securing the second position in the forgetting metric on the 20-split benchmark. DualLoRA+ outperforms the state-of-the-art scheme InfLoRA by $2.58\%$, $7.14\%$, and $4.96\%$ in terms of average accuracy metric across the 5-split, 10-split, and 20-split settings. In addition, in terms of the forgetting metric, DualLoRA+ shows improvements over InfLoRA by $1.95\%$, $4.14\%$, and $3.23\%$, respectively, in the same settings. 
\\

\noindent \textbf{CIFAR100 and Tiny-ImageNet.} DualLoRA steadily shows strong performance in these two datasets, achieving the best performance in average accuracy compared to prior existing schemes on CIFAR100 and Tiny-ImageNet benchmarks while slightly underperforming CodaPrompt on Tiny-ImageNet in forgetting metrics. DualLoRA+ consistently demonstrates extraordinary performance in average accuracy, outperforming CodaPrompt (the best scheme in these baselines) by $5.17\%$, $2.07\%$ and $1.04 \%$, respectively, and also demonstrating an advantage in the forgetting metric on 10-split CIFAR100 and 10-split Tiny-ImageNet. To give more insights, we report the average accuracy computed with different numbers of learned tasks, as illustrated in Fig.~\ref{acc_vs_time}(a) and ~\ref{acc_vs_time}(b) As shown in the figures, DualLoRA and DualLoRA+ outperform other baselines in different stages of continual learning. Furthermore, DualLoRA+ demonstrates robust resistance to forgetting as the number of learned tasks increases, suggesting the potential of DualLoRA+ fine-tuning foundational models across a wider range of tasks without significant forgetting.

\begin{figure}[t] 
    \centering
	  \subfloat[FLOPs]{
        \includegraphics[width=0.5\linewidth]
        {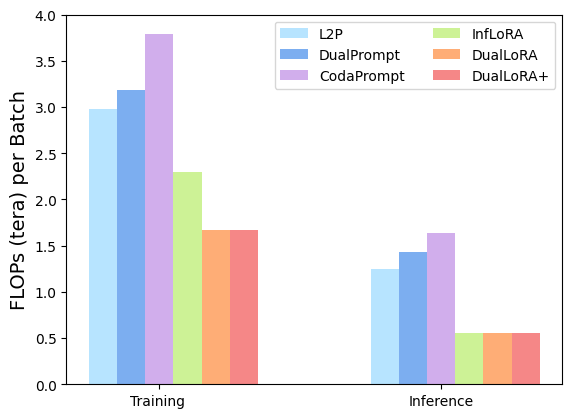}}
        \subfloat[Inference Time]{
        \includegraphics[width=0.496\linewidth]
        {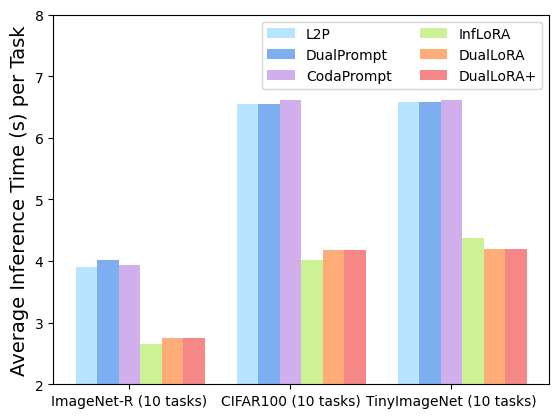}}
	\caption{Figure (a) demonstrates the approximated average FLOPs during training and
inference on each batch of data points. Figure (b) demonstrates the actual average running time for
different schemes to perform inference on a task.}
\label{more_results} 
\end{figure}

\begin{table*}[t]
\caption{\textbf{O} stands for orthogonal adapter, \textbf{R} stands for residual adapter, and \textbf{Task ID} stands for giving true task ID during inference.}
\centering  
\begin{tabular}{lcccccc}
\bottomrule[1pt]
\label{table3}
\multirow{2}{*}{\textbf{Method} }    & 
  \multicolumn{2}{c}{\textbf{10-Split ImageNet-R}} &\multicolumn{2}{c}{\textbf{10-Split CIFAR100}}&\multicolumn{2}{c}{\textbf{10-Split TinyImageNet}}\\
   &    ACC($\uparrow$) & FT($\downarrow$) & ACC($\uparrow$) & FT($\downarrow$) &ACC($\uparrow$) & FT($\downarrow$)\\
\hline  
LoRA    &61.85 & 26.0  & 73.03 &20.26&67.69 &23.70\\
LoRA + O  &74.11  &4.65  &84.03 &5.67 &83.92 &7.74\\
LoRA + O + R    &74.60 &4.12  &86.65 &3.96 &85.61 &4.30\\
DualLoRA    &76.23   & 3.67 &89.13 &5.08 &86.42 &3.87 \\
DualLoRA+    &81.17  &2.04  &90.94 &3.20 &87.74 &2.45 \\
\hline
DualLoRA + Task ID   & \textbf{87.66} &\textbf{0.46}  &\textbf{94.39} &\textbf{ 1.05} &\textbf{95.71} &\textbf{0.84}\\
\toprule[1pt]
\end{tabular}
\end{table*}

\subsection{Ablation Study}
We perform additional experiments to confirm the effectiveness of various subroutines within the DualLoRA scheme. To be specific, we implement three variants of DualLoRA: (1) \textbf{LoRA + O} stands for only using the orthogonal adapter; (2) \textbf{LoRA + O + R} stands for running DualLoRA with orthogonal and residual adapters but performing no task identity prediction; (3)\textbf{ LoRA + O + R + Task ID} assumes knowing true task identities. As illustrated in Table~\ref{table3}, the orthogonal adapter significantly improves the performance of LoRA while the residual adapter further enhances both average accuracy and forgetting metrics. Moreover, DualLoRA and DualLoRA+ further improve the performance by using individual features and average features to perform task identity prediction. The final configuration, where true task identities are utilized, exhibits superior performance in both average accuracy and forgetting, highlighting the crucial role of task prediction.

\subsection{Computation and Inference Time}
To better compare DualLoRA with other baselines in terms of computation, we report the number of floating point operations (FLOPs) during the training and inference phases in Figure~\ref{more_results}(a) and the average inference time on a task in Figure~\ref{more_results}(b) According to the results in the figure, InfLoRA has the lowest FLOPs during inference, while DualLoRA has the lowest FLOPs during training because InfLoRA requires a double forward pass. 
During inference, InfLoRA and DualLoRA have similar inference time across different datasets, which is less than 50\% inference time compared to the prompt-based CL schemes. Details on computing FLOPs are provided in Appendix B.2.

\begin{table}[t]
\caption{Metrics computed from experiments on ImageNet-R (10 tasks) using various pretrained ViTs beyond ImageNet-21k. $\overline{\textbf{ACC}}$ denotes the average of \textbf{ACC} in every timestep.} 
\centering
\begin{tabular}{clcc}
\label{table6}
&\textbf{Method} &\textbf{ACC}(\%) &$\overline{\textbf{ACC}}$(\%) \\
\hline  
\multirow{5}{*}{\textbf{ImageNet-1k}}&L2P &60.23 & 67.86 \\
&DualPrompt   &67.45 & 72.48\\
&CodaPrompt  &73.26 &79.49 \\
&InfLoRA  & 75.58 & 81.92 \\
&DualLoRA  & \textbf{77.28} &\textbf{82.63}\\
\hline  
\multirow{5}{*}{\textbf{SAM-1k}}&L2P & 52.71& 58.24  \\
&DualPrompt  &56.57 &63.23 \\
&CodaPrompt  &61.13 & 70.79\\
&InfLoRA   & 64.62 &73.66\\
&DualLoRA  &\textbf{66.44} &\textbf{74.70}\\
\toprule[1pt]
\end{tabular}
\end{table}

\subsection{Varying the Pre-Trained Models}
To evaluate the consistency of DualLoRA's performance, we conduct experiments using the ViT-B/16 model, pretrained on ImageNet-1k through both supervised learning and the unsupervised SAM framework \citep{sam}. To represent the average performance of all schemes during the continual learning process, we use $\overline{\text{ACC}} = \frac{1}{T}\sum_{t = 1}^{T} \text{ACC}_{t}$ as an additional metric in the table. As shown, all schemes experience performance degradation when using a ViT model pretrained on unsupervised datasets. Nonetheless, DualLoRA consistently outperforms other schemes in terms of average accuracy.

\section{Conclusion}
We introduce DualLoRA, a novel low-rank adaptation scheme for vision transformers (ViTs) that integrates orthogonal and residual adapters, operating in parallel with pre-trained weights. This structure achieves a balance between stability and plasticity in continual learning through a dynamic memory mechanism that leverages subspaces from previously learned tasks. Furthermore, we develop a task identity prediction scheme based on core bases extracted from each learned task to enhance DualLoRA’s performance. Extensive experiments demonstrate that DualLoRA outperforms state-of-the-art continual learning methods across multiple benchmarks while requiring fewer computational resources. We see this work as an important step toward developing more efficient and effective continual learning paradigms for foundational models.

{
    \small
    \bibliographystyle{ieeenat_fullname}
    \bibliography{main}

\begin{thebibliography}{33}
\providecommand{\natexlab}[1]{#1}
\providecommand{\url}[1]{\texttt{#1}}
\expandafter\ifx\csname urlstyle\endcsname\relax
  \providecommand{\doi}[1]{doi: #1}\else
  \providecommand{\doi}{doi: \begingroup \urlstyle{rm}\Url}\fi

\bibitem[Cha et~al.(2021)Cha, Lee, and Shin]{co2l}
Hyuntak Cha, Jaeho Lee, and Jinwoo Shin.
\newblock Co2l: Contrastive continual learning.
\newblock In \emph{Proceedings of the IEEE/CVF International conference on computer vision}, pages 9516--9525, 2021.

\bibitem[Cha et~al.(2025)Cha, Chen, and Vikalo]{cha2025task}
Seohyeon Cha, Huancheng Chen, and Haris Vikalo.
\newblock Task-agnostic federated continual learning via replay-free gradient projection.
\newblock \emph{arXiv preprint arXiv:2509.21606}, 2025.

\bibitem[Deng et~al.(2021)Deng, Chen, Hao, Wang, and Heng]{FSGPM}
Danruo Deng, Guangyong Chen, Jianye Hao, Qiong Wang, and Pheng-Ann Heng.
\newblock Flattening sharpness for dynamic gradient projection memory benefits continual learning.
\newblock \emph{Advances in Neural Information Processing Systems}, 34:\penalty0 18710--18721, 2021.

\bibitem[Farajtabar et~al.(2020)Farajtabar, Azizan, Mott, and Li]{OGD}
Mehrdad Farajtabar, Navid Azizan, Alex Mott, and Ang Li.
\newblock Orthogonal gradient descent for continual learning.
\newblock In \emph{International Conference on Artificial Intelligence and Statistics}, pages 3762--3773. PMLR, 2020.

\bibitem[Gao et~al.(2023)Gao, Zhao, Sun, Xi, Zhang, Ghanem, and Zhang]{gao2023unified}
Qiankun Gao, Chen Zhao, Yifan Sun, Teng Xi, Gang Zhang, Bernard Ghanem, and Jian Zhang.
\newblock A unified continual learning framework with general parameter-efficient tuning.
\newblock In \emph{Proceedings of the IEEE/CVF International Conference on Computer Vision}, pages 11483--11493, 2023.

\bibitem[Gao et~al.(2024)Gao, Dong, He, Wang, and Gong]{gao2024beyond}
Xinyuan Gao, Songlin Dong, Yuhang He, Qiang Wang, and Yihong Gong.
\newblock Beyond prompt learning: Continual adapter for efficient rehearsal-free continual learning.
\newblock In \emph{European Conference on Computer Vision}, pages 89--106. Springer, 2024.

\bibitem[He et~al.(2025)He, Duan, and Zhu]{CL-LoRA}
Jiangpeng He, Zhihao Duan, and Fengqing Zhu.
\newblock Cl-lora: Continual low-rank adaptation for rehearsal-free class-incremental learning.
\newblock In \emph{Proceedings of the Computer Vision and Pattern Recognition Conference}, pages 30534--30544, 2025.

\bibitem[Hendrycks et~al.(2021)Hendrycks, Basart, Mu, Kadavath, Wang, Dorundo, Desai, Zhu, Parajuli, Guo, et~al.]{imagenetr}
Dan Hendrycks, Steven Basart, Norman Mu, Saurav Kadavath, Frank Wang, Evan Dorundo, Rahul Desai, Tyler Zhu, Samyak Parajuli, Mike Guo, et~al.
\newblock The many faces of robustness: A critical analysis of out-of-distribution generalization.
\newblock In \emph{Proceedings of the IEEE/CVF international conference on computer vision}, pages 8340--8349, 2021.

\bibitem[Hu et~al.(2021)Hu, Shen, Wallis, Allen-Zhu, Li, Wang, Wang, and Chen]{lora}
Edward~J Hu, Yelong Shen, Phillip Wallis, Zeyuan Allen-Zhu, Yuanzhi Li, Shean Wang, Lu Wang, and Weizhu Chen.
\newblock Lora: Low-rank adaptation of large language models.
\newblock \emph{arXiv preprint arXiv:2106.09685}, 2021.

\bibitem[Kirillov et~al.(2023)Kirillov, Mintun, Ravi, Mao, Rolland, Gustafson, Xiao, Whitehead, Berg, Lo, et~al.]{sam}
Alexander Kirillov, Eric Mintun, Nikhila Ravi, Hanzi Mao, Chloe Rolland, Laura Gustafson, Tete Xiao, Spencer Whitehead, Alexander~C Berg, Wan-Yen Lo, et~al.
\newblock Segment anything.
\newblock In \emph{Proceedings of the IEEE/CVF International Conference on Computer Vision}, pages 4015--4026, 2023.

\bibitem[Kong et~al.(2022)Kong, Liu, Wang, and Tao]{adns}
Yajing Kong, Liu Liu, Zhen Wang, and Dacheng Tao.
\newblock Balancing stability and plasticity through advanced null space in continual learning.
\newblock In \emph{European Conference on Computer Vision}, pages 219--236. Springer, 2022.

\bibitem[Lester et~al.(2021)Lester, Al-Rfou, and Constant]{lester2021power}
Brian Lester, Rami Al-Rfou, and Noah Constant.
\newblock The power of scale for parameter-efficient prompt tuning.
\newblock \emph{arXiv preprint arXiv:2104.08691}, 2021.

\bibitem[Liang and Li(2023)]{dualgpm}
Yan-Shuo Liang and Wu-Jun Li.
\newblock Adaptive plasticity improvement for continual learning.
\newblock In \emph{Proceedings of the IEEE/CVF Conference on Computer Vision and Pattern Recognition}, pages 7816--7825, 2023.

\bibitem[Liang and Li(2024)]{inflora}
Yan-Shuo Liang and Wu-Jun Li.
\newblock Inflora: Interference-free low-rank adaptation for continual learning.
\newblock \emph{arXiv preprint arXiv:2404.00228}, 2024.

\bibitem[Lin et~al.(2022)Lin, Yang, Fan, and Zhang]{trgp}
Sen Lin, Li Yang, Deliang Fan, and Junshan Zhang.
\newblock Trgp: Trust region gradient projection for continual learning.
\newblock \emph{arXiv preprint arXiv:2202.02931}, 2022.

\bibitem[Loo et~al.(2020)Loo, Swaroop, and Turner]{loo2020generalized}
Noel Loo, Siddharth Swaroop, and Richard~E Turner.
\newblock Generalized variational continual learning.
\newblock \emph{arXiv preprint arXiv:2011.12328}, 2020.

\bibitem[Lu et~al.(2024)Lu, Zhang, Cheng, Xing, Wang, Wang, and Zhang]{lu2024visual}
Yue Lu, Shizhou Zhang, De Cheng, Yinghui Xing, Nannan Wang, Peng Wang, and Yanning Zhang.
\newblock Visual prompt tuning in null space for continual learning.
\newblock \emph{arXiv preprint arXiv:2406.05658}, 2024.

\bibitem[Luo et~al.(2026)Luo, Zhou, Zhang, Wan, Wei, and Zhang]{keeplora}
Mao-Lin Luo, Zi-Hao Zhou, Yi-Lin Zhang, Yuanyu Wan, Tong Wei, and Min-Ling Zhang.
\newblock Keeplora: Continual learning with residual gradient adaptation.
\newblock \emph{arXiv preprint arXiv:2601.19659}, 2026.

\bibitem[McDonnell et~al.(2023)McDonnell, Gong, Parvaneh, Abbasnejad, and Van~den Hengel]{mcdonnell2023ranpac}
Mark~D McDonnell, Dong Gong, Amin Parvaneh, Ehsan Abbasnejad, and Anton Van~den Hengel.
\newblock Ranpac: Random projections and pre-trained models for continual learning.
\newblock \emph{Advances in Neural Information Processing Systems}, 36:\penalty0 12022--12053, 2023.

\bibitem[Qiao et~al.(2023)Qiao, Tan, Chen, Qu, Peng, Xie, et~al.]{pgp}
Jingyang Qiao, Xin Tan, Chengwei Chen, Yanyun Qu, Yong Peng, Yuan Xie, et~al.
\newblock Prompt gradient projection for continual learning.
\newblock In \emph{The Twelfth International Conference on Learning Representations}, 2023.

\bibitem[Qiu et~al.(2025)Qiu, Zhang, Qiao, Guan, Zhang, and Nie]{splitlora}
Haomiao Qiu, Miao Zhang, Ziyue Qiao, Weili Guan, Min Zhang, and Liqiang Nie.
\newblock Splitlora: Balancing stability and plasticity in continual learning through gradient space splitting.
\newblock \emph{arXiv preprint arXiv:2505.22370}, 2025.

\bibitem[Saha and Roy(2023)]{sgp}
Gobinda Saha and Kaushik Roy.
\newblock Continual learning with scaled gradient projection.
\newblock In \emph{Proceedings of the AAAI Conference on Artificial Intelligence}, pages 9677--9685, 2023.

\bibitem[Saha et~al.(2021)Saha, Garg, and Roy]{gpm}
Gobinda Saha, Isha Garg, and Kaushik Roy.
\newblock Gradient projection memory for continual learning.
\newblock \emph{arXiv preprint arXiv:2103.09762}, 2021.

\bibitem[Smith et~al.(2023)Smith, Karlinsky, Gutta, Cascante-Bonilla, Kim, Arbelle, Panda, Feris, and Kira]{coda}
James~Seale Smith, Leonid Karlinsky, Vyshnavi Gutta, Paola Cascante-Bonilla, Donghyun Kim, Assaf Arbelle, Rameswar Panda, Rogerio Feris, and Zsolt Kira.
\newblock Coda-prompt: Continual decomposed attention-based prompting for rehearsal-free continual learning.
\newblock In \emph{Proceedings of the IEEE/CVF Conference on Computer Vision and Pattern Recognition}, pages 11909--11919, 2023.

\bibitem[Wang et~al.(2024)Wang, Zhang, Su, and Zhu]{wang2024comprehensive}
Liyuan Wang, Xingxing Zhang, Hang Su, and Jun Zhu.
\newblock A comprehensive survey of continual learning: Theory, method and application.
\newblock \emph{IEEE Transactions on Pattern Analysis and Machine Intelligence}, 2024.

\bibitem[Wang et~al.(2021)Wang, Li, Sun, and Xu]{NSCL}
Shipeng Wang, Xiaorong Li, Jian Sun, and Zongben Xu.
\newblock Training networks in null space of feature covariance for continual learning.
\newblock In \emph{Proceedings of the IEEE/CVF conference on Computer Vision and Pattern Recognition}, pages 184--193, 2021.

\bibitem[Wang et~al.(2022{\natexlab{a}})Wang, Huang, and Hong]{Sprompt}
Yabin Wang, Zhiwu Huang, and Xiaopeng Hong.
\newblock S-prompts learning with pre-trained transformers: An occam’s razor for domain incremental learning.
\newblock \emph{Advances in Neural Information Processing Systems}, 35:\penalty0 5682--5695, 2022{\natexlab{a}}.

\bibitem[Wang et~al.(2022{\natexlab{b}})Wang, Zhang, Ebrahimi, Sun, Zhang, Lee, Ren, Su, Perot, Dy, et~al.]{dualprompt}
Zifeng Wang, Zizhao Zhang, Sayna Ebrahimi, Ruoxi Sun, Han Zhang, Chen-Yu Lee, Xiaoqi Ren, Guolong Su, Vincent Perot, Jennifer Dy, et~al.
\newblock Dualprompt: Complementary prompting for rehearsal-free continual learning.
\newblock In \emph{European Conference on Computer Vision}, pages 631--648. Springer, 2022{\natexlab{b}}.

\bibitem[Wang et~al.(2022{\natexlab{c}})Wang, Zhang, Lee, Zhang, Sun, Ren, Su, Perot, Dy, and Pfister]{l2p}
Zifeng Wang, Zizhao Zhang, Chen-Yu Lee, Han Zhang, Ruoxi Sun, Xiaoqi Ren, Guolong Su, Vincent Perot, Jennifer Dy, and Tomas Pfister.
\newblock Learning to prompt for continual learning.
\newblock In \emph{Proceedings of the IEEE/CVF Conference on Computer Vision and Pattern Recognition}, pages 139--149, 2022{\natexlab{c}}.

\bibitem[Wu et~al.(2025)Wu, Piao, Huang, Wang, Li, Pfister, Meng, Ma, and Wei]{sdlora}
Yichen Wu, Hongming Piao, Long-Kai Huang, Renzhen Wang, Wanhua Li, Hanspeter Pfister, Deyu Meng, Kede Ma, and Ying Wei.
\newblock Sd-lora: Scalable decoupled low-rank adaptation for class incremental learning.
\newblock \emph{arXiv preprint arXiv:2501.13198}, 2025.

\bibitem[Zhao et~al.(2023)Zhao, Zhang, Tan, Liu, Qu, Xie, and Ma]{SD}
Zhen Zhao, Zhizhong Zhang, Xin Tan, Jun Liu, Yanyun Qu, Yuan Xie, and Lizhuang Ma.
\newblock Rethinking gradient projection continual learning: Stability/plasticity feature space decoupling.
\newblock In \emph{Proceedings of the IEEE/CVF Conference on Computer Vision and Pattern Recognition}, pages 3718--3727, 2023.

\bibitem[Zhou et~al.(2024)Zhou, Sun, and et~al.]{zhou2024expandable}
Da-Wei Zhou, Hai-Long Sun, and et al.
\newblock Expandable subspace ensemble for pre-trained model-based class-incremental learning.
\newblock In \emph{Proceedings of the IEEE/CVF Conference on Computer Vision and Pattern Recognition}, pages 23554--23564, 2024.

\bibitem[Zhu et~al.(2025)Zhu, Zhang, Dong, and Koniusz]{bilora}
Hao Zhu, Yifei Zhang, Junhao Dong, and Piotr Koniusz.
\newblock Bilora: almost-orthogonal parameter spaces for continual learning.
\newblock In \emph{Proceedings of the IEEE/CVF Conference on Computer Vision and Pattern Recognition}, pages 25613--25622, 2025.

\end{thebibliography}
}

\end{document}


\maketitle
\section*{Appendix A.1: Approximate Change of Output Activation}
In main paper, we introduce the approximate output change after orthogonal adapter $\mathbf{O}^{(l)}_{t}$ fine-tuning on the $(t+1)$-th task. In what follows we drop the layer superscript $(l)$ and let $\mathcal{S}(\cdot)$ denote \textbf{softmax} operation.
\begin{equation}
    \begin{aligned}
    &\Delta \mathbf{h}_{t+1} = \mathbf{h}_{t+1} - \mathbf{h}_{t} \nonumber \\
    &=  \mathcal{S} \left( \frac{\mathbf{Q}\left(\mathbf{K}_{t+1}\right)^{\top}}{\sqrt{d}} \right) \cdot \mathbf{V}_{t+1} - \mathcal{S} \left( \frac{\mathbf{Q}\left(\mathbf{K}_{t}\right)^{\top}}{\sqrt{d}} \right) \cdot \mathbf{V}_{t} \nonumber \\
    &= \mathcal{S} \left( \frac{\mathbf{Q} (\mathbf{W}_{t+1}^{k})^{\top} \mathbf{a}^{\top}}{\sqrt{d}} \right) \cdot \mathbf{V}_{t+1} - \mathcal{S} \left( \frac{\mathbf{Q} (\mathbf{W}_{t}^{k})^{\top} \mathbf{a}^{\top}}{\sqrt{d}} \right) \cdot \mathbf{V}_{t} \nonumber \\
    &= \mathcal{S} \left( \frac{\mathbf{Q} (\mathbf{W}_{t}^{k} + \Delta \mathbf{O}_{t+1}^{k})^{\top} \mathbf{a}^{\top}}{\sqrt{d}} \right) \cdot \mathbf{a} (\mathbf{W}_{t}^{v} + \Delta \mathbf{O}_{t+1}^{v}) \\
    &\quad - \mathcal{S} \left( \frac{\mathbf{Q} (\mathbf{W}_{t}^{k})^{\top} \mathbf{a}^{\top}}{\sqrt{d}} \right) \cdot \mathbf{a} \mathbf{W}_{t}^{v} \nonumber \\  
    &= \mathcal{A} \cdot \mathbf{a} \mathbf{W}_{t}^{v}  + \mathcal{B} \cdot \mathbf{a}\Delta \mathbf{O}_{t+1}^{v} \nonumber, \\
\end{aligned}
\end{equation}
where
\begin{equation}
    \mathcal{A} = \mathcal{S} \left( \frac{\mathbf{Q} (\mathbf{W}_{t}^{k} +\Delta \mathbf{O}_{t+1}^{k})^{\top} \mathbf{a}^{\top}}{\sqrt{d}} \right) -  \mathcal{S} \left( \frac{\mathbf{Q} (\mathbf{W}_{t}^{k})^{\top} \mathbf{a}^{\top}}{\sqrt{d}} \right),
\end{equation}
\begin{equation}
    \mathcal{B} = \mathcal{S} \left( \frac{\mathbf{Q} (\mathbf{W}_{t}^{k} +\Delta \mathbf{O}_{t+1}^{k})^{\top} \mathbf{a}^{\top}}{\sqrt{d}} \right).
\end{equation}
Then, considering
\begin{equation}
\mathcal{A} 
= \mathcal{S} \left( \mathbf{z} + \Delta\mathbf{z}\right) -  \mathcal{S} \left( \mathbf{z}\right),\\
\end{equation}
where $ \mathbf{z} = \frac{\mathbf{Q} (\mathbf{W}_{t}^{k})^T \mathbf{a}^T}{\sqrt{d}}, \Delta  \mathbf{z} =  \frac{\mathbf{Q} (\Delta \mathbf{O}_{t+1}^{k})^T \mathbf{a}^T}{\sqrt{d}}$. Since $\left \Vert \Delta \mathbf{O}_{t+1}^{k} \right \Vert$ is proportional to learning rate $\eta$, by selecting a small learning rate, we can get $\left \Vert \Delta \mathbf{z}\right \Vert \ll \left \Vert \mathbf{z}\right \Vert$. Then considering 
\begin{equation}
    \begin{aligned}
    \mathcal{A}_{i} &= \mathcal{S} \left( \mathbf{z} + \Delta\mathbf{z}\right)_{i} - \mathcal{S} \left( \mathbf{z} \right)_{i}=   \sum_{j=1}^{N}\frac{\partial S_{i}}{\partial \mathbf{z}_{j}} \Delta \mathbf{z}_{j},
    \end{aligned}
\end{equation}
where $S_{i} = \mathcal{S} \left( \mathbf{z} \right)_{i} < 1$ is the $i$-th component of $ \mathcal{S} \left( \mathbf{z} \right)$; $\Delta \mathbf{z}_{j}$ denotes the $j$-th component of $\Delta \mathbf{z}$. Since $\frac{\partial S_{i}}{\partial \mathbf{z}_{j}} = S_{i}(1- S_{j})$ if $i = j$ and $\frac{\partial S_{i}}{\partial \mathbf{z}_{j}} = - S_{i}S_{j}$, otherwise, we can obtain:
\begin{equation}
\begin{aligned}
        \mathcal{A}_{i} &=  S_{i}(1- S_{i}) \Delta\mathbf{z}_{i} -S_{i}\sum_{j \not =i} S_{j} \Delta \mathbf{z}_{j}\\
        &= S_{i}\Delta\mathbf{z}_{i} - S_{i}\sum_{j = 1}^{N} S_{j} \Delta \mathbf{z}_{j} \\
        & \leq S_{i}\Delta\mathbf{z}_{i} - S_{i}\min_{j} (\Delta \mathbf{z}_{j} ) \\
        &= \left( S_{i} - \frac{\min_{j} (\Delta \mathbf{z}_{j})}{\Delta\mathbf{z}_{i}}\right)\Delta\mathbf{z}_{i}.
\end{aligned}
\end{equation}
Similarly, we can get 
\begin{equation}
     \left( S_{i} - \frac{\max_{j} (\Delta \mathbf{z}_{j})}{\Delta\mathbf{z}_{i}}\right)\leq  \frac{\mathcal{A}_{i} }{\Delta\mathbf{z}_{i}}\leq \left( S_{i} - \frac{\min_{j} (\Delta \mathbf{z}_{j})}{\Delta\mathbf{z}_{i}}\right).
\end{equation}
Let $\mathcal{A}_{i} = \gamma_{i} \Delta\mathbf{z}_{i}$ such that $\left( S_{i} - \frac{\max_{j} (\Delta \mathbf{z}_{j})}{\Delta\mathbf{z}_{i}}\right) \leq \gamma_{i} \leq  \left( S_{i} - \frac{\min_{j} (\Delta \mathbf{z}_{j})}{\Delta\mathbf{z}_{i}}\right)$, then we have $\mathcal{A} = \Gamma \cdot \Delta \mathbf{z}$ where $\Gamma = \text{diag}\{\gamma_{1}, \dots, \gamma_{N}\}$. Then we refocus on $\mathcal{B}$ such that
\begin{equation}
\begin{aligned}
    \mathcal{B} &= \mathcal{S} \left( \frac{\mathbf{Q} (\mathbf{W}_{t}^{k} +\Delta \mathbf{O}_{t+1}^{k})^{\top} \mathbf{a}^{\top}}{\sqrt{d}} \right)\\
    &= \mathcal{S} \left(\mathbf{z} + \Delta \mathbf{z}\right).
\end{aligned}
\end{equation}
For each component,
\begin{equation}
\begin{aligned}
    \mathcal{B}_{i} &= \mathcal{S} \left(\mathbf{z} + \Delta \mathbf{z}\right)_{i}\\
    &= \mathcal{S} \left(\mathbf{z}\right)_{i} + \mathcal{A}_{i}\\
    &= \mathcal{S} \left(\mathbf{z}\right)_{i} + S_{i}\Delta\mathbf{z}_{i} - S_{i}\sum_{j = 1}^{N} S_{j} \Delta \mathbf{z}_{j}.\\
\end{aligned}
\end{equation}
Since $\Delta \mathbf{z}$ and $\Delta \mathbf{O}_{t+1}^{v}$ are small, we then get
\begin{equation}
\begin{aligned}
    \mathcal{B} \cdot \mathbf{a}\Delta \mathbf{O}_{t+1}^{v} &\approx  \mathcal{S} \left(\mathbf{z}\right) \cdot \mathbf{a}\Delta \mathbf{O}_{t+1}^{v} \\
    & = \mathcal{S} \left( \frac{\mathbf{Q} (\mathbf{W}_{t}^{k})^{\top} \mathbf{a}^{\top}}{\sqrt{d}} \right) \cdot \mathbf{a}\Delta \mathbf{O}_{t+1}^{v}\\
    &=  \mathcal{S} \left( \frac{\mathbf{Q} \left(\mathbf{K}\right)^{\top}}{\sqrt{d}} \right) \cdot \mathbf{a} \Delta \mathbf{O}_{t+1}^{v}.
\end{aligned}
\end{equation}
Combining $\mathcal{A} \cdot \mathbf{a} \mathbf{W}_{t}^{v}$ and $\mathcal{B} \cdot \mathbf{a}\Delta \mathbf{O}_{t+1}^{v}$, we get
\begin{equation}
\begin{aligned}
     \Delta \mathbf{h}^{(l)} &\approx  \Gamma \cdot \frac{\mathbf{Q}^{(l)} \left(\mathbf{a}^{(l)} \Delta \mathbf{O}_{t+1}^{k} \right)^{\top}}{\sqrt{d}}  \cdot \mathbf{V}^{(l)}\\
    &\quad + \mathcal{S} \left( \frac{\mathbf{Q}^{(l)} \left(\mathbf{K}^{(l)}\right)^{\top}}{\sqrt{d}} \right) \cdot \mathbf{a}^{(l)} \Delta \mathbf{O}_{t+1}^{v}.
\end{aligned}
\end{equation}

$\hfill \blacksquare$

\section*{Appendix A.2: Major Change of $\Delta \mathbf{h}_{1}^{(l)}$}
We aim to use gradient projection to \textbf{mitigate} forgetting, similar to prior gradient projection schemes such as GPM \citep{gpm} and InfLoRA \citep{inflora}. In practice, strictly performing orthogonal gradient descent can preserve previously learned features but also restrict the subspace for fine-tuning on new tasks. Therefore, these schemes typically set a threshold value, such as $\epsilon=0.95$, to extract a subset of significant core bases, selecting the top $95\%$ of singular values to construct the feature subspaces. Although these CL schemes cannot perfectly preserve previously learned representations, a significant reduction in changes can effectively mitigate forgetting.

Similarly, our DualLoRA extracts only the feature subspace relevant to the class token and performs gradient projection to reduce the change in $\Delta \mathbf{h}_{1}^{(l)}$. We acknowledge that we cannot keep the class token unchanged but aim to prevent major changes to it. In the main paper, we simplified the error which can be generalized as
\begin{equation}
\small
\begin{aligned}
      & \Delta \mathbf{h}^{(l)} \approx  \Gamma \cdot \frac{(\mathbf{Q}^{(l)} + \Delta\mathbf{Q}^{(l)} )\left(\Delta \mathbf{O}_{t+1}^{k} \right)^{\top} (\mathbf{a}^{(l)}  + \Delta\mathbf{a}^{(l)})^{\top}}{\sqrt{d}}  \cdot  \mathbf{V}^{(l)} \\
   & + \Gamma \cdot \frac{(\mathbf{Q}^{(l)} + \Delta\mathbf{Q}^{(l)} )\left(\Delta \mathbf{O}_{t+1}^{k} \right)^{\top} (\mathbf{a}^{(l)}  + \Delta\mathbf{a}^{(l)})^{\top}}{\sqrt{d}}  \cdot  \Delta \mathbf{V}^{(l)}\\
   & + \text{softmax} \left( \frac{(\mathbf{Q}^{(l)} + \Delta\mathbf{Q}^{(l)} )\left(\mathbf{K}^{(l)} + \Delta\mathbf{K}^{(l)}\right)^{\top}}{\sqrt{d}} \right) \cdot \mathbf{a}^{(l)}  \Delta \mathbf{O}_{t+1}^{v}\\
   & + \text{softmax} \left( \frac{(\mathbf{Q}^{(l)} + \Delta\mathbf{Q}^{(l)} )\left(\mathbf{K}^{(l)} + \Delta\mathbf{K}^{(l)}\right)^{\top}}{\sqrt{d}} \right) \cdot \Delta \mathbf{a}^{(l)} \Delta \mathbf{O}_{t+1}^{v},
\end{aligned}
\end{equation}
where 
\begin{equation}
\small
    \Delta\mathbf{a}^{(l)} = \Delta \mathbf{h}^{(l-1)} \cdot \mathbf{W}_{\text{FFN}} ,
\end{equation}
\begin{equation}
\small
    \Delta\mathbf{Q}^{(l)} = \Delta \mathbf{h}^{(l-1)} \cdot \mathbf{W}_{\text{FFN}} \cdot \mathbf{W}^{q},
\end{equation}
\begin{equation}
\small
\Delta\mathbf{V}^{(l)} = \Delta \mathbf{h}^{(l-1)} \cdot \mathbf{W}_{\text{FFN}} \cdot (\mathbf{W}^{v} + \mathbf{O}^{v} ) 
\end{equation}
\begin{equation}
\small
\Delta\mathbf{K}^{(l)} = \Delta \mathbf{h}^{(l-1)} \cdot \mathbf{W}_{\text{FFN}} (\mathbf{W}^{v} + \mathbf{O}^{k}). 
\end{equation}
In the above formula, $\mathbf{W}_{\text{FFN}}$ denotes feedforward network. Assuming $\Delta \mathbf{h}_{1}^{(l-1)} = \mathbf{0}$, then $\Delta\mathbf{a}_{1}^{(l)}  = \Delta \mathbf{Q}_{1}^{(l)} = \Delta \mathbf{K}_{1}^{(l)} = \Delta \mathbf{V}_{1}^{(l)}  = \mathbf{0}$. Since there are two terms on the right-hand side, we consider 
\begin{equation}
\small
     \Delta \mathbf{h}^{(l)} \approx \mathcal{K} + \mathcal{V},
\end{equation}
where $\mathcal{K}$ is relevant to the updates $\Delta \mathbf{O}_{t+1}^{k}$ while $\mathcal{V}$ is relevant to the updates $\Delta \mathbf{O}_{t+1}^{v}$. For clarity, we omit the superscript $l$ and subscript $t+1$ and recall that $\Delta \mathbf{h} \in \mathbb{R}^{n \times d}$, $\mathbf{a} \in \mathbb{R}^{n \times d}$, $\Delta \mathbf{O}^{k} \in \mathbb{R}^{d \times d}$, $\Delta \mathbf{O}^{v} \in \mathbb{R}^{d \times d}$, $ \mathbf{V} \in \mathbb{R}^{n \times d}$, $ \mathbf{Q} \in \mathbb{R}^{n \times d}$, $ \mathbf{K} \in \mathbb{R}^{n \times d}$. Let consider the first term $\mathcal{K}$, (ignoring the constants), 
\begin{equation}
\small
\begin{aligned}
    &(\mathbf{Q} + \Delta \mathbf{Q} )\left(\Delta \mathbf{O}^{k}\right)^{\top} (\mathbf{a} + \Delta \mathbf{a})^{\top} \cdot (\mathbf{V} + \Delta \mathbf{V})  = \\
    &\left[ \begin{array}{c}
	(\mathbf{Q}_{1} + \mathbf{0}) \cdot  (\Delta \mathbf{O}^{k})^{\top} \cdot  (\mathbf{a} + \Delta \mathbf{a})^{\top} \\
	\vdots\\
	(\mathbf{Q}_{n} + \Delta \mathbf{Q}_{n})\cdot  (\Delta \mathbf{O}^{k})^{\top} \cdot  (\mathbf{a} + \Delta \mathbf{a})^{\top} \\
	\end{array}
	\right] \left( \mathbf{V}^{(l)} + \Delta \mathbf{V}^{(l)}\right),
\end{aligned}
\end{equation}
where $\mathbf{Q}_{i} \in \mathbb{R}^{1\times d}$ is the $i$-th row of $\mathbf{Q}$. We randomly select $m$ samples for computing the layer-wise features set $\mathbf{k}^{(l)}$ including $m$ varying $\mathbf{Q}_{1}$ vectors and extracting the core bases $\boldsymbol{\Phi}^{k}$. By projection, we update $\Delta \mathbf{O}^{k}$ by
\begin{equation}
\small
    (\Delta \mathbf{O}^{k})^{\top} \leftarrow  \left(\mathbf{I} - (\boldsymbol{\Phi}^{k})^{\top}\boldsymbol{\Phi}^{k}\right)\Delta (\mathbf{O}^{k})^{\top}.
\end{equation}
Ideally, we can get 
\begin{equation}
\small
    \mathbf{Q}_{1} \cdot  \left(\mathbf{I} - (\boldsymbol{\Phi}^{k})^{\top}\boldsymbol{\Phi}^{k}\right)\Delta (\mathbf{O}^{k})^{\top} \approx \mathbf{0} \cdot  \Delta (\mathbf{O}^{k})^{\top} = \mathbf{0}.
\end{equation}
Therefore, we can constraint $\mathcal{K}_{1}$, the first row of $\mathcal{K}$, in a small value close to zero. Similarly, let consider the value of $\mathcal{V}$. Let $\mathbf{X} = \text{softmax}\left(\frac{(\mathbf{Q} + \Delta\mathbf{Q} )\left(\mathbf{K}+ \Delta\mathbf{K}\right)^{\top}}{\sqrt{d}}\right)\in \mathbb{R}^{n \times n}$, we can ignore the higher order infinitesimal since $\Delta \mathbf{Q}(\Delta \mathbf{K})^{\top} \propto \Delta \mathbf{h}^{(l-1)}(\Delta \mathbf{h}^{(l-1)})^{\top}$, then
\begin{equation}
\small
    \mathbf{X}  =  \text{softmax}\left( \frac{\mathbf{Q}\mathbf{K}^{\top}}{\sqrt{d}} + \frac{\Delta \mathbf{Q}\mathbf{K}^{\top}}{\sqrt{d}} +\frac{\mathbf{Q}\Delta\mathbf{K}^{\top}}{\sqrt{d}}\right).
\end{equation}
Therefore,
\begin{equation}
\small
\begin{aligned}
    \mathbf{X}_{1} &= \text{softmax}\left(\frac{1}{\sqrt{d}} \mathbf{Q}_{1}\mathbf{K}^{\top} + \frac{1}{\sqrt{d}}\Delta \mathbf{Q}_{1}\mathbf{K}^{\top}  + \frac{1}{\sqrt{d}}\mathbf{Q}_{1}\Delta \mathbf{K}^{\top}\right) \\
    & = \text{softmax}\left(\frac{1}{\sqrt{d}} \mathbf{Q}_{1}\mathbf{K}^{\top}  + \frac{1}{\sqrt{d}}\mathbf{Q}_{1}\Delta \mathbf{K}^{\top}\right).
\end{aligned}
\end{equation}
According to derivative of softmax function, we get
\begin{equation}
\small
    \mathbf{X}_{1} = \text{softmax}\left(\frac{1}{\sqrt{d}} \mathbf{Q}_{1}\mathbf{K}^{\top}\right) + \mathbf{H}_{1} \frac{1}{\sqrt{d}}\mathbf{Q}_{1}\Delta \mathbf{K}^{\top},
\end{equation}
where $\mathbf{H}$ is the Jacobian matrix defined as
\begin{equation}
\small
    \mathbf{H} = \left[
	\begin{array}{cccc}
	\mathbf{p}_{1}(1 - \mathbf{p}_{1}) & -\mathbf{p}_{1}\mathbf{p}_{2}&\cdots& -\mathbf{p}_{1}\mathbf{p}_{n}\\
	-\mathbf{p}_{2}\mathbf{p}_{1}&\mathbf{p}_{2}(1 - \mathbf{p}_{2})\\
	\vdots&&\ddots&\vdots\\
	-\mathbf{p}_{n}\mathbf{p}_{1}&-\mathbf{p}_{n}\mathbf{p}_{2}&\cdots&\mathbf{p}_{n}(1 - \mathbf{p}_{n})\\
	\end{array}
	\right] ,
\end{equation}
where $\mathbf{p}_{i}$ is the $i$-th component of $\text{softmax}\left(\frac{1}{\sqrt{d}} \mathbf{Q}_{1}\mathbf{K}^{\top}\right)$. Therefore, the first row of $\mathcal{V}$ can be found as:
\begin{equation}
\small
\begin{aligned}
        \mathcal{V}_{1} & = \mathbf{X}_{1} \cdot \left( \mathbf{a} + \Delta \mathbf{a}\right) \Delta \mathbf{O}^{v} \\
        & =  \text{softmax}\left(\frac{1}{\sqrt{d}} \mathbf{Q}_{1}\mathbf{K}^{\top}\right) \mathbf{a} \Delta \mathbf{O}^{v} \\
        & \quad + \text{softmax}\left(\frac{1}{\sqrt{d}} \mathbf{Q}_{1}\mathbf{K}^{\top}\right) \Delta \mathbf{a} \Delta \mathbf{O}^{v} \\
        & \quad +  \mathbf{H}_{1} \frac{1}{\sqrt{d}}\mathbf{Q}_{1}\Delta \mathbf{K}^{\top}\mathbf{a}  \Delta \mathbf{O}^{v} +  \mathbf{H}_{1} \frac{1}{\sqrt{d}}\mathbf{Q}_{1}\Delta \mathbf{K}^{\top}\Delta \mathbf{a}  \Delta \mathbf{O}^{v},
\end{aligned}
\end{equation}
where $\mathbf{X}_{1} \in \mathbb{R}^{1 \times n}$, $\mathbf{a}\in \mathbb{R}^{n \times d}$ and $\mathbf{O}^{v}\in \mathbb{R}^{d \times d}$. Similarly, we can eliminate the fourth term since higher  order infinitesimal $\Delta\mathbf{K}^{\top} \Delta \mathbf{a} \propto \Delta \mathbf{h}^{(l-1)}(\Delta \mathbf{h}^{(l-1)})^{\top} \approx \mathbf{0}$. According to our methodology, we collect feature set $\mathbf{s}^{(l)}$ and project $\Delta \mathbf{O}^{v}$ to constraint
\begin{equation}
\small
    \text{softmax}\left(\frac{1}{\sqrt{d}} \mathbf{Q}_{1}\mathbf{K}^{\top}\right) \mathbf{a} \Delta \mathbf{O}^{v} = \mathbf{0}.
\end{equation}
In summary, if we guarantee the change of activation from last layer $\Delta \mathbf{h}^{(l-1)} \approx \mathbf{0}$, then
\begin{equation}
\small
\begin{aligned}
    \Delta \mathbf{h}_{1}^{(l)} \approx \mathcal{K}_{1} + \mathcal{V}_{1} & =  \text{softmax}\left(\frac{1}{\sqrt{d}} \mathbf{Q}_{1}\mathbf{K}^{\top}\right) \Delta \mathbf{a} \Delta \mathbf{O}^{v} \\
    & \quad +  \mathbf{H}_{1} \frac{1}{\sqrt{d}}\mathbf{Q}_{1}\Delta \mathbf{K}^{\top} \mathbf{a}  \Delta \mathbf{O}^{v}.
\end{aligned}
\end{equation}
Although $\Delta \mathbf{a}$, $\Delta \mathbf{O}^{v}$ and $\Delta \mathbf{K}$ are not the same infinitesimal, but $\Delta \mathbf{a} \propto \eta$, $\Delta \mathbf{O}^{v}\propto \eta$ and  $\Delta \mathbf{K}\propto \eta$, where $\eta$ is the learning rate. Therefore,
\begin{equation}
    \Delta \mathbf{h}_{1}^{(l)} \propto \eta^{2}. 
\end{equation}
If $\eta$ is sufficiently small, we can have a very small change on $\Delta \mathbf{h}_{1}^{(l)}$. 
\section*{Appendix B.1: Experimental Details}
In this section, we report hyper-parameters used in different methods in Table.~\ref{table1}. In the table, we use $\eta$ to denote learning rate; $p$ denotes total number of prompts in the prompt pool; $e$ denotes length of prompts and $k$ denotes the number of prompts needed to match input images; $l$ denotes the number of layer expanding parameters. For DualPrompt, $e_E$ and $e_G$ denote the length of E-prompts and G-prompt, respectively; $l_{E}$ and $l_{G}$ denotes the number of layers instructed by E-prompt and G-prompt. $r$ denotes rank of LoRA parameters; $\epsilon$ denotes the accumulated singular value for extracting bases in SVD.

\begin{table*}[h]
\caption{List of Hyper-parameters used in different schemes.   } 
\centering
\small  
\begin{tabular}{ll}
\bottomrule[1pt]
\label{table4}
\textbf{Method}   &  \textbf{Hyper-parameters} \\
\hline  
L2P   &$\eta:$  0.03 (ImageNet-R, CIFAR100, Tiny-ImageNet)  \\
      &$p:$ 30 (ImageNet-R, CIFAR100, Tiny-ImageNet) 
 \\
     &$e:$  20 (ImageNet-R, CIFAR100, Tiny-ImageNet) 
 \\
      &$k:$  5 (ImageNet-R, CIFAR100, Tiny-ImageNet) 
 \\
       &$l:$ 1 (ImageNet-R, CIFAR100, Tiny-ImageNet) 
 \\
DualPrompt   &$\eta:$  0.03 (ImageNet-R, CIFAR100, Tiny-ImageNet)  \\
       &$p:$ 5 (5 Tasks), 10 (10 Tasks), 20 (20 Tasks) \\ 
     &$e_{E}:$  20 (ImageNet-R, CIFAR100, Tiny-ImageNet) 
 \\
      &$e_{G}:$  6 (ImageNet-R, CIFAR100, Tiny-ImageNet) 
 \\
      &$k:$  5 (ImageNet-R, CIFAR100, Tiny-ImageNet) 
 \\
        &$l_{E}:$  3 (ImageNet-R, CIFAR100, Tiny-ImageNet) 
 \\
        &$l_{G}:$  2 (ImageNet-R, CIFAR100, Tiny-ImageNet) 
 \\
PGP   &$\eta:$  0.05 (ImageNet-R, Tiny-ImageNet), 0.03 (CIFAR100)  \\
       &$p:$ 5 (5 Tasks), 10 (10 Tasks), 20 (20 Tasks)  \\
     &$e_{E}:$  20 (ImageNet-R, CIFAR100, Tiny-ImageNet) 
 \\
      &$e_{G}:$  6 (ImageNet-R, CIFAR100, Tiny-ImageNet) 
 \\
      &$k:$  5 (ImageNet-R, CIFAR100, Tiny-ImageNet) 
 \\
        &$l_{E}:$ 3 (ImageNet-R, CIFAR100, Tiny-ImageNet) 
 \\
        &$l_{G}:$  2 (ImageNet-R, CIFAR100, Tiny-ImageNet) 
 \\
SPrompt   &$\eta:$ 0.0005 (ImageNet-R, CIFAR100, Tiny-ImageNet)  \\
       &$p:$ 5 (5 Tasks), 10 (10 Tasks), 20 (20 Tasks)  
 \\
      &$e:$  10 (ImageNet-R, CIFAR100, Tiny-ImageNet) 
 \\
         &$k:$ 5 (ImageNet-R, CIFAR100, Tiny-ImageNet) 
 \\
        &$l:$ 1 (ImageNet-R, CIFAR100, Tiny-ImageNet) 
 \\
CodaPrompt   &$\eta:$ 0.0005 (ImageNet-R, CIFAR100, Tiny-ImageNet)  \\
       &$p:$  100 (ImageNet-R, CIFAR100, Tiny-ImageNet) 
 \\
      &$e:$  8 (ImageNet-R, CIFAR100, Tiny-ImageNet) 
 \\
       &$k:$  5 (ImageNet-R, CIFAR100, Tiny-ImageNet) 
 \\
        &$l:$  5 (ImageNet-R, CIFAR100, Tiny-ImageNet) 
 \\
 InfLoRA   &$\eta:$ 0.0005 (ImageNet-R, CIFAR100, Tiny-ImageNet)  \\
    &$r:$ 10 (ImageNet-R, CIFAR100, Tiny-ImageNet)  \\
    &$\epsilon:$ 0.95 ( CIFAR100), 0.98 (ImageNet-R, Tiny-ImageNet)  \\
        &$l:$ 12 ( CIFAR100), 0.98 (ImageNet-R, Tiny-ImageNet)  \\

 DualLoRA   &$\eta:$ 0.0005 (ImageNet-R, CIFAR100, Tiny-ImageNet)  \\
    &$r:$ 10 (ImageNet-R, CIFAR100, Tiny-ImageNet)  \\
    &$\epsilon:$ 0.95 (ImageNet-R, CIFAR100, Tiny-ImageNet)  \\
    &$m:$ 200 (5 Tasks), 150 (10 Tasks), 100 (20 Tasks)  \\
    &$\lambda:$ 2 (CIFAR100, 10-Split ImageNet-R), 1.5 (5-Split ImageNet-R, Tiny-ImageNet)\\
    &$\quad$ 1.2 (20-Split ImageNet-R)  \\
    &$l:$ 12 ( CIFAR100), 0.98 (ImageNet-R, Tiny-ImageNet)  \\

\toprule[1pt]
\end{tabular}
\end{table*}
\section*{Appendix B.2: FLOPs Computation}
In this section, we present formulas to estimate floating point operations (FLOPs) to facilitate the comparison of computational demands across different methods. For simplicity, we concentrate on matrix multiplication and disregard operations with minor computational costs, such as addition, dropout, normalization and computing activation. For two matrices $\mathbf{A} \in \mathbb{R}^{m \times n}$ and $\mathbf{B} \in \mathbb{R}^{n \times p}$, the FLOPs for multiplication can be found as $2mnp$.
\subsection*{Forward Pass in ViT}
\textbf{Multi-Head Attention.} Computation in multi-head attention block involves in 
\begin{itemize}
    \item Obtaining $\mathbf{Q}$, $\mathbf{K}$ and $\mathbf{V}$ by multiplying $\mathbf{x}$ with $\mathbf{W}^{q}$, $\mathbf{W}^{k}$ and $\mathbf{W}^{v}$. Since $\mathbf{x} \in \mathbb{R}^{b \times n \times d}$, $\mathbf{W}^{q},\mathbf{W}^{k} ,\mathbf{W}^{v} \in \mathbb{R}^{d \times d}$, FLOPs in this step can be found as:
    \begin{equation}
        \textbf{FLOPs} = 3 \cdot 2 \cdot b n d^{2} = 6bnd^{2},
    \end{equation}
    where $b$ is batch size, $n$ is length of sequence and $d$ denotes the dimension of embedding.
    \item Computing attention score $S$ by multiplying  $\mathbf{Q}$ and $\mathbf{K}$. FLOPs in this step can be found as :
    \begin{equation}
        \textbf{FLOPs} = 2  b n^{2} d.
    \end{equation}
    \item Computing output signal $\mathbf{h}$ by multiplying  $\mathbf{S}$ and $\mathbf{V}$, FLOPs in this step can be found as :
    \begin{equation}
        \textbf{FLOPs} = 2  b n^{2} d.
    \end{equation}
    \item Linear projection on the same shape with $\mathbf{h}$, FLOPs in this step can be found as:
    \begin{equation}
         \textbf{FLOPs} = 2  b n d^{2}.
    \end{equation}
\end{itemize}
Overall, the total FLOPs needed each MHA block for one batch is $8bnd^{2} + 4bn^{2}d$.
\\

\noindent \textbf{The Feedforward Network (FFN).} There are two linear projection for decoding and encoding output signals $\mathbf{h}$ in FFN. Suppose the ratio is set to $4$, the FLOPs can be found as:
\begin{equation}
    \textbf{FLOPs} = 2 \cdot 8bnd^{2} = 16bnd^{2}
\end{equation}
Since ViT model does not need to compute word embedding in the output layer for each block, we do not consider computation in this part.
Suppose there are $L$ blocks in the ViT model, the total FLOPs needed in the forward pass of ViT can be compuated as $\textbf{FLOPs} = L(24bnd^{2} + 4bn^{2}d)$.

\subsection*{Backward Pass in ViT}
The FLOPs required for the backward pass are simply double those needed for the forward pass with the same model. There, the FLOPs needed for backward pass can be found as:
\begin{equation}
    \textbf{FLOPs} = 2L(24bnd^{2} + 4bn^{2}d)
\end{equation}

\subsection*{Singular Value Decomposition}
SVD of a matrix $\mathbf{A} \in \mathbb{R}^{d \times m}$ typically involves two phases: (1) reduction to bidiagonal form and (2) performing the decomposition using the Golub-Kahan algorithm. The second phase is iterative, making it difficult to determine the exact FLOPs required. However, we focus on the FLOPs needed for the first phase, as it dominates the overall computational cost. According to the textbook \citep{trefethen2022numerical}, the FLOPs for the first phase can be found as 
\begin{equation}
    \textbf{FLOPs} = 2dm^{2}+11m^{3}.
\end{equation}

\subsection*{Forward Pass in LoRA module}
LoRA parameters are assigned parallel to the pre-trained weights $\mathbf{W}^{k}$ and $\mathbf{W}^{v}$ causing additional FLOPs to forward the signals. Since LoRA parameters for each per-trained weight consist of $\mathbf{A} \in \mathbb{R}^{d \times r}$ and $\mathbf{B}^{r \times d}$ , the additional FLOPs can be found as
\begin{equation}
    \textbf{FLOPs} = L \cdot 2 \cdot 2 \cdot 2bndr = 8Lndr
\end{equation}

\subsection*{Forward Pass in DualLoRA module}
Compared to original LoRA, DualLoRA assigns an additional residual adapter parallel to the value weight $\mathbf{W}^{v}$. Therefore, the addintional FLOPs in DualLoRA can be found as
\begin{equation}
    \textbf{FLOPs} = L \cdot (2+1) \cdot 2 \cdot 2bndr = 12Lndr
\end{equation}

\subsection*{Overall FLOPs in L2P}
During the training and inference phases, L2P \citep{l2p} first forwards image tokens to the original pre-trained encoder to obtain the key needed for matching prompt vectors in the prompt pool. Then, the selected prompt vectors are concatenated with the image tokens and forwarded into the encoder again. Therefore, the forward pass computation needs to be counted twice. For simplicity, we ignore the computation needed in the minimizing problem to select the top $k$ prompt vectors because the FLOPs is depending on the optimization algorithm. Therefore, we can getting a lower bound for  the overall FLOPs in L2P. For training phase, the overall FLOPs for a batch data can be found as
\begin{equation}
\begin{aligned}
    \textbf{FLOPs} & \geq 3 \cdot L\left(24b(n+ke)d^{2} + 4b(n+ke)^{2}d\right) \\
    & \quad + L\left(24bnd^{2} + 4bn^{2}d\right),
\end{aligned}
\end{equation}
where $e$ is the length of prompt vectors. And the overall FLOPs for the inference phase can be found as:
\begin{equation}
\begin{aligned}
    \textbf{FLOPs} & \geq   L\left(24b(n+ke)d^{2} + 4b(n+ke)^{2}d\right) \\
    & \quad + L\left(24bnd^{2} + 4bn^{2}d\right)
\end{aligned}
\end{equation}

\subsection*{Overall FLOPs in DualPrompt}
DualPrompt \citep{dualprompt} has a workflow similar to L2P but processes prompt vectors in only a subset of layers. For simplicity, we ignore the matching algorithm in DualPrompt and estimate the lower bound of the overall FLOPs. Therefore, the overall FLOPs can be calculated as follows:
\begin{equation}
\label{dualprompt1}
\begin{aligned}
    &\textbf{FLOPs} \geq 72Lb(n+\frac{2e_{G}}{L} + \frac{3ke_{E}}{L})d^{2} \\
    & + 12Lb(n+\frac{2e_{G}}{L} + \frac{3ke_{E}}{L})^{2}d + L \left( 24bnd^{2} + 4bn^{2}d\right),
\end{aligned}
\end{equation}
where $e_{G}$ is the length of G-prompts and $e_{E}$ is the length of E-prompts. Similarly, for inference phase:
\begin{equation}
\label{dualprompt2}
\begin{aligned}
    \textbf{FLOPs} &\geq  L\left(24b(n+e_{G} + ke_{E})d^{2} +  24bnd^{2} + 4bn^{2}d\right)\\
    &\quad + 4Lb(n+\frac{2e_{G}}{L} + \frac{3ke_{E}}{L})^{2}d.
\end{aligned}
\end{equation}

\subsection*{Overall FLOPs in CODAPrompt}
CodaPrompt \citep{coda} requires significantly more FLOPs for optimizing the prompt keys and prompt pool. However, quantifying the exact number of FLOPs is challenging due to its dependence on the minimization algorithm. In addition to this computation, CodaPrompt increases the size of the matching prompt vectors for the first $l$ layers. Therefore, the lower bound for the overall FLOPs in the training phase can be estimated as follows:
\begin{equation}
\begin{aligned}
     \textbf{FLOPs} \geq &3 \cdot l\left(24b(n+\frac{1+l}{2}ke)d^{2} + 4b(n+\frac{1+l}{2}ke)^{2}d\right) \\
     &+ 3 \cdot (L-l)\left(24b(n+ lke)d^{2} + 4b(n+lke)^{2}d\right) \\
     &+ L\left(24bnd^{2} + 4bn^{2}d\right).
\end{aligned}
\end{equation}
 And the overall FLOPs for the inference phase can be found as:
\begin{equation}
\begin{aligned}
     \textbf{FLOPs} \geq & l\left(24b(n+\frac{1+l}{2}ke)d^{2} + 4b(n+\frac{1+l}{2}ke)^{2}d\right) \\
     &+  (L-l)\left(24b(n+ lke)d^{2} + 4b(n+lke)^{2}d\right) \\
     & + L\left(24bnd^{2} + 4bn^{2}d\right),
\end{aligned}
\end{equation}

\subsection*{Overall FLOPs in InfLoRA}
To obtain the gradient subspace for each task, InfLoRA \citep{inflora} requires forwarding the entire training dataset through the model and performing SVD on the collected average gradient. Additionally, InfLoRA forwards the data through the model once more for parameter updates. Therefore, there is also twice FLOPs in forward pass in the training phase but only one forward pass in the inference phase. Combing the FLOPs in forward pass, LoRA pass and SVD, the overall FLOPs in the training phase can be found as
\begin{equation}
    \textbf{FLOPs} = 4L\left(24bnd^{2} + 4bn^{2}d\right) + 8Lndr + 13Ld^{3},
\end{equation}
And the overall FLOPs for the inference phase can be found as:
\begin{equation}
    \textbf{FLOPs} = L\left(24bnd^{2} + 4bn^{2}d\right) + 8Lndr,
\end{equation}

\subsection*{Overall FLOPs in DualLoRA}
Since DualLoRA does not require forwarding the training data twice during training, the overall FLOPs in the training phase can be found as
\begin{equation}
\begin{aligned}
    \textbf{FLOPs} & = 3L\left(24bnd^{2} + 4bn^{2}d\right) + 12Lndr \\
    &\quad + L(2dm^{2}+11m^{3}),
\end{aligned}
\end{equation}
And the overall FLOPs for the inference phase can be found as:
\begin{equation}
    \textbf{FLOPs} = L\left(24bnd^{2} + 4bn^{2}d\right) + 12Lndr,
\end{equation}
{
    \small
    \bibliographystyle{ieeenat_fullname}
    \bibliography{main}
}